%% file: main.tex
\documentclass[sigconf]{acmart}

\usepackage[normalem]{ulem}

\usepackage{booktabs} 
\usepackage{amssymb}
\usepackage{enumerate}
\usepackage{color}
\usepackage{graphicx}
\usepackage[titlenumbered,ruled]{algorithm2e}
\usepackage{array} 
\usepackage[group-separator={,},group-minimum-digits=4]{siunitx} 
\usepackage{booktabs} 
\usepackage{multirow}
\usepackage[inline]{enumitem} 
\usepackage[noend]{algorithmic}
\usepackage{footnote}
\usepackage{tikz}
\makesavenoteenv{tabular}
\makesavenoteenv{table}
\usepackage{natbib}
\usepackage{MnSymbol}
\usetikzlibrary{arrows,patterns}
\usepackage{multicol}
\usepackage[colorinlistoftodos,prependcaption,textsize=footnotesize,color=green]{todonotes}
\makeatletter
\usepackage{regexpatch}
\xpatchcmd{\@todo}{\setkeys{todonotes}{#1}}{\setkeys{todonotes}{inline,#1}}{}{}
\makeatother

\usepackage[font=small,labelfont=bf]{caption}

\setcopyright{rightsretained}

\copyrightyear{2017} 
\acmYear{2017} 
\setcopyright{acmlicensed}
\acmConference{CIKM'17 }{November 6--10, 2017}{Singapore, Singapore}\acmPrice{15.00}\acmDOI{10.1145/3132847.3133041}
\acmISBN{978-1-4503-4918-5/17/11}

\begin{document}
\title{Taxonomy Induction Using Hypernym Subsequences}

\newcommand{\forceindent}{\leavevmode{\parindent=0.6em\indent}}

\newcommand*{\new}{\textcolor{violet}}

\renewcommand{\shortauthors}{Anonymized}

\author{Amit Gupta}
\affiliation{%
  \institution{EPFL, Lausanne, Switzerland}
}
\email{amit.gupta@epfl.ch}

\author{R\'emi Lebret}
\affiliation{%
  \institution{EPFL, Lausanne, Switzerland}
}
\email{remi.lebret@epfl.ch}

\author{Hamza Harkous}
\affiliation{%
  \institution{EPFL, Lausanne, Switzerland}
}
\email{hamza.harkous@epfl.ch }

\author{Karl Aberer}
\affiliation{%
  \institution{EPFL, Lausanne, Switzerland}
}
\email{karl.aberer@epfl.ch}

\begin{abstract}
\input{abstract}
\end{abstract}

%
%





\maketitle

\input{intro}

\input{taxonomy}

\input{semeval}

\input{realeval}

\input{related}

\input{conclusions}
\paragraph*{Acknowledgements} This work is supported by a Sinergia Grant by the Swiss National Science Foundation (SNF 147609). The authors would like to thank Marius Pa\c{s}ca for helpful discussions.
\bibliographystyle{ACM-Reference-Format}
\bibliography{sigproc} 

\end{document}

%% file: abstract.tex
We propose a novel, semi-supervised approach towards domain taxonomy induction from an input vocabulary of seed terms. Unlike all previous approaches, which typically extract direct hypernym \textit{edges} for terms, our approach utilizes a novel probabilistic framework to extract hypernym \textit{subsequences}. Taxonomy induction from extracted subsequences is cast as an instance of the minimum-cost flow problem on a carefully designed directed graph. Through experiments, we demonstrate that our approach outperforms state-of-the-art taxonomy induction approaches across four languages. Importantly, we also show that our approach is robust to the presence of noise in the input vocabulary. To the best of our knowledge, this robustness has not been empirically proven in any previous approach.

%% file: intro.tex
\section{Introduction}
\label{sec:intro}

\paragraph{\textbf{Motivation.}}Lexical semantic knowledge in the form of term taxonomies has been beneficial in a variety of NLP tasks, including inference, textual entailment, question answering and information extraction~\citep{biemann2005ontology}. This widespread utility of taxonomies has led to multiple large-scale manual efforts towards taxonomy induction, such as WordNet~\citep{miller1995wordnet} and Cyc ~\citep{lenat1995cyc}. However, such manually constructed taxonomies suffer from low coverage~\citep{hovy2009toward} and are unavailable for specific domains or languages. Therefore, in recent years, there has been substantial interest in extending existing taxonomies automatically or building new ones ~\citep{snow2006semantic,yang2009metric,kozareva2010semi,velardi2013ontolearn,task17semeval2015,task13semeval2016}.

Approaches towards automated taxonomy induction consist of two main stages: 
\begin{enumerate}
\item
\textbf{extraction of hypernymy relations} (i.e., ``is-a" relations between a term and its hypernym such as \textit{apple}$\rightarrow$\textit{fruit})
\item
\textbf{ structured organization of terms into a taxonomy}, i.e., a coherent tree-like hierarchy. 
\end{enumerate}

Extraction of hypernymy relations has been relatively well-studied in previous works. Its approaches can be classified into two main categories: \textit{Distributional} and \textit{Pattern-based} approaches.

\textit{Distributional} approaches use clustering to extract hypernymy relations from structured or unstructured text. Such approaches draw primarily on the distributional hypothesis~\citep{harris1954distributional}, which states that semantically similar terms appear in similar contexts. The main advantage of distributional approaches is that they can discover relations not directly expressed in the text. 
\begin{figure*}[t]
 \centering        \includegraphics[width=0.8\linewidth]{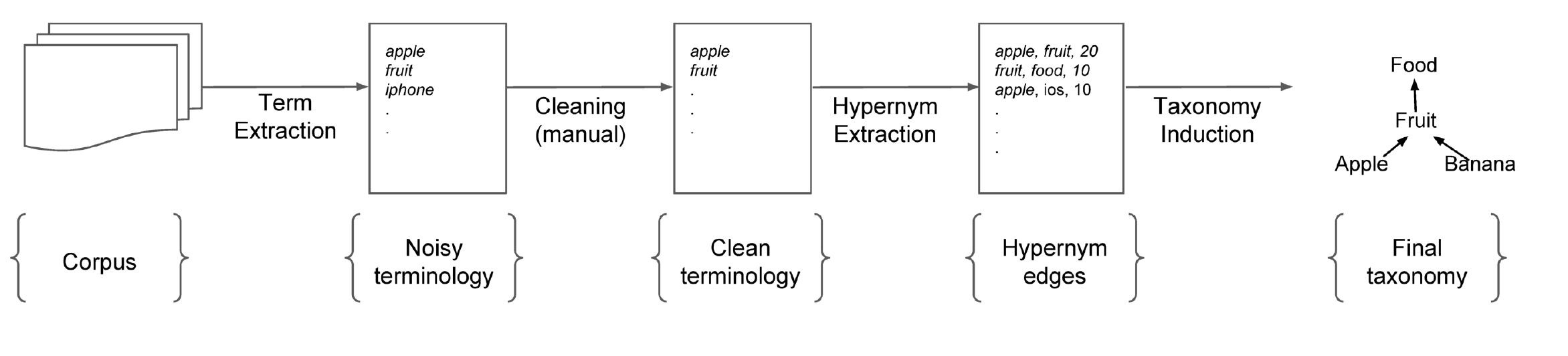}
        \caption{Traditional process for taxonomy induction from a domain-specific corpus~\cite{velardi2013ontolearn}.}
        \label{fig:process} 
\end{figure*}
In contrast, \textit{Pattern-based} approaches utilize pre-defined rules or lexico-syntactic patterns to extract terms and hypernymy relations from text~\citep{hearst1992automatic,oakes2005using}. Patterns are either chosen manually~\citep{hearst1992automatic,kozareva2008semantic} or learnt automatically via bootstrapping~\citep{snow2004learning}.  
Pattern-based approaches usually result in higher accuracies. However, unlike the distributional approaches, which are fully unsupervised, they require a set of seed surface patterns to initiate the extraction process. 

Early work on the second stage of taxonomy induction, namely the structured organization of terms into a taxonomy, focused on extending existing partial taxonomies such as WordNet by inserting missing terms at appropriate positions~\cite{widdows2003unsupervised,snow2006semantic,yang2009metric}. Another line of work focused on taxonomy induction from Wikipedia by exploiting the semi-structured nature of the Wikipedia category network~\cite{suchanek2007yago,ponzetto2008wikitaxonomy,ponzetto2011taxonomy,nastase2010wikinet,flati2016multiwibi,guptarevisiting}.

Subsequent approaches to taxonomy induction focused on building lexical taxonomies entirely \textit{from scratch}, i.e., from a domain corpus or the Web~\cite{kozareva2010semi,navigli2011graph,velardi2013ontolearn,bansal2014structured,alfarone2015unsupervised,panchenko2016taxi}. 

Automated taxonomy induction from scratch is preferred because it can be used over arbitrary domains, including highly specific or technical domains, such as Finance or Artificial Intelligence~\cite{navigli2011graph}. Such domains are usually under-represented in existing taxonomic resources. For example, WordNet is limited to the most frequent and the most important nouns, adjectives, verbs, and adverbs~\cite{gurevych2010expert,nakashole2012patty}. Similarly, Wikipedia is limited to popular entities~\cite{kliegr2014linked}, and its utility is further diminished by slowed growth~\cite{suh2009singularity}.

Past approaches to taxonomy induction from scratch either assume the availability of a clean input vocabulary~\cite{panchenko2016taxi} or employ a time-consuming manual cleaning step over a noisy input vocabulary~\cite{velardi2013ontolearn}. For example, Figure~\ref{fig:process} shows the pipeline of a typical taxonomy induction approach from a domain corpus~\cite{velardi2013ontolearn}. An initial noisy vocabulary is automatically extracted from the domain corpus using a term extraction tool, such as \textit{TermExtractor}~\citep{sclano2007termextractor}, and is further cleaned manually to produce the final vocabulary. This requirement severely limits the applicability of such approaches in an automated setting because clean vocabularies are usually unavailable for specific domains.

To handle these limitations, we designed our approach to induce a taxonomy directly from a noisy input vocabulary. Consequently, it is the first work to fully automate the taxonomy induction process for arbitrary domains.

\paragraph{\textbf{Contributions.}} 
In this paper, we present a novel, semi-supervised approach for building lexical taxonomies given an input vocabulary of (potentially noisy) seed terms.
We leverage the existing work on hypernymy relations extraction and focus on the second stage, i.e. the organization of terms into a taxonomy. Our main contributions are as follows:

\begin{itemize}
\item 
We propose a novel probabilistic framework for extracting longer hypernym subsequences from hypernymy relations, as well as a novel minimum-cost flow based optimization framework for inducing a tree-like taxonomy from a noisy hypernym graph.
\item
We empirically show that our approach outperforms state-of-the-art taxonomy induction approaches across four different languages, while achieving $>$32\% relative improvement in F1-measure over the Food domain.
\item
We demonstrate that our subsequence-based model is robust to the presence of noisy terms in the input vocabulary, and achieves a 65\% relative improvement in precision over an edge-based model while maintaining similar coverage. To the best of our knowledge, this is the first approach towards taxonomy induction from a noisy input vocabulary. 
\end{itemize}

The rest of the paper is organized as follows. In Section~\ref{sec:tax}, we describe our taxonomy induction approach. In Section~\ref{sec:eval}, we discuss our experiments and performance results. In Section~\ref{sec:related}, we discuss related work. We conclude in Section~\ref{sec:conc}.

%% file: taxonomy.tex
\section{Taxonomy Induction}
\label{sec:tax}

Given a potentially-noisy vocabulary\footnote{In this work, we use terminology and vocabulary interchangeably.} of seed terms as an input, we define our goal as inducing a taxonomy consisting of these seed terms (and possibly other terms). 
This taxonomy is a directed acyclic graph with terms as the nodes and the edges indicating a hypernymy relationship between the terms. 
For our task, we assume the availability of a database of \textit{candidate} hypernymy relations. 
Multiple such resources have been compiled and made available publicly over the years. A prominent example of such a resource is WebIsA~\citep{seitner2016large}, a collection of more than 400 million hypernymy relations for English, extracted from the CommonCrawl web corpus using lexico-syntactic patterns. However, such resources come with a considerable number of noisy candidate hypernyms, typically containing a mixture of relations such as hyponymy, meronymy, synonymy and co-hyponymy. For example, WebIsA has more than 12,000 hypernyms for the term \textit{apple}, including noisy hypernyms such as \textit{orange}, \textit{everyone} and \textit{smartphone}. A sample set of candidate hypernyms and their occurrence frequencies for the term \textit{apple} taken from WebIsA is shown in Table~\ref{tab:apple_hyp}. 

Our approach to taxonomy induction consists of three main steps:
\begin{enumerate}
\item
extracting hypernym subsequences for the given seed terms (Section~\ref{sec:hyp}),
\item
aggregating the extracted subsequences into an initial hypernym graph (Section~\ref{sec:agg}),
\item
pruning the hypernym graph using a minimum-cost flow approach to induce the final taxonomy (Section~\ref{sec:flow}). 
\end{enumerate}

\subsection{Hypernym Subsequences Extraction}
\label{sec:hyp}

Unsupervised or semi-supervised approaches to taxonomy induction typically aim to extract \mbox{\textbf{single hypernym edges}} among terms from noisy candidate hypernyms \citep{kozareva2010semi,panchenko2016taxi}. 
In contrast, our approach consists of extracting  \mbox{\textbf{hypernym subsequences}} (where a subsequence is a series of one or more individual hypernym edges).

\begin{table}[t]
\centering
\footnotesize
\begin{tabular}{cc}
    \toprule
    Candidate hypernym & Frequency \\
    \midrule
    company & 5536  \\
    fruit & 3898\\
    apple & 2119\\
    vegetable & 928 \\ 
    orange & 797\\
    tech company & 619 \\
    brand & 463 \\
    hardware company & 460 \\
    technology company & 427 \\
    food & 370 \\
    \bottomrule
  \end{tabular}
  \caption{Candidate hypernyms for the term \textit{apple}.}%
 \label{tab:apple_hyp}
\end{table}

To motivate this, we first note that Table~\ref{tab:apple_hyp} includes hypernyms of \textit{apple} at different levels of generality, such as \textit{fruit} and \textit{food}. 
In fact, we observe this pattern in the candidate hypernyms of most terms. 
This suggests that we can leverage such information to not only extract the direct hypernyms of \textit{apple}, but to also extract longer hypernym subsequences, such as \textit{apple}$\rightarrow$\textit{fruit}$\rightarrow$\textit{food}. 

\begin{figure}[t]
\centering
\includegraphics[width=0.7\linewidth]{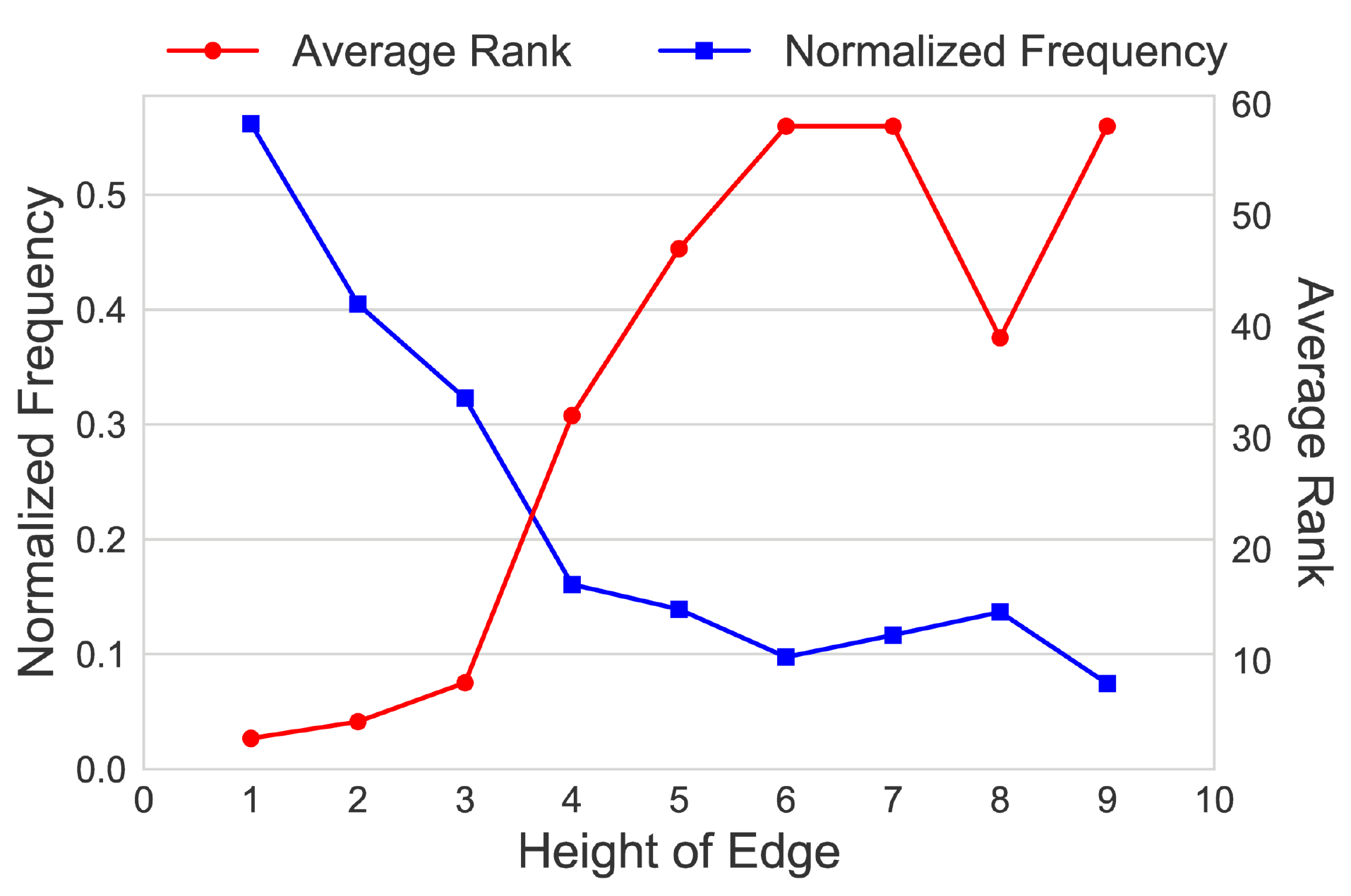}
\caption{Average rank and normalized frequency of WordNet edges vs. height of edge. }
\label{fig:rank_vs_height}
\end{figure}

This becomes even more important given the result by~\citet{velardi2013ontolearn}, who demonstrated that hypernym extraction becomes increasingly erroneous as the generality of terms increases, mainly due to the increase in term ambiguity. To further support this hypothesis, we perform an experiment where we first randomly sample 100 paths from Wordnet. For each edge $a$$\rightarrow$$b$ in a sampled path, we plot the normalized frequency\footnote{Normalization is performed by dividing frequency counts by the maximum.} of ``$b$ as a candidate hypernym for $a$'' against the height of the edge, where frequencies are computed using lexico-syntactic patterns (cf. Table~\ref{tab:apple_hyp}). We also plot the average rank of $b$ among candidate hypernyms of $a$, where candidate hypernyms are ranked by their normalized frequencies in a decreasing order. Results of this experiment are shown in Figure~\ref{fig:rank_vs_height}. Since edges in WordNet are assumed to be ground truth, it is desired that they have a higher normalized frequency and lower ranks.  This small-scale experiment demonstrates that as the height of the edge increases, the normalized frequencies decrease whereas the average ranks increase. Therefore, the accuracy of patterns-based hypernymy detection decreases for more general terms that appear higher in generalization paths. Hence, for such terms, it makes sense to not solely base the hypernym selection on a noisy set of candidate hypernyms. 
We can potentially improve the accuracy of selected hypernyms for general terms (such as \textit{fruit}) by relying on extracted subsequences starting from more specific terms (such as \textit{apple}). Those subsequences would be evidenced by the less-noisy candidate hypernyms of the specific terms.

In sum, extracting hypernym subsequences is both \textit{possible} and potentially \textit{beneficial}. The remainder of this section describes our model that realizes this intuition.
\begin{figure}[t]
\includegraphics[width=0.33\linewidth]{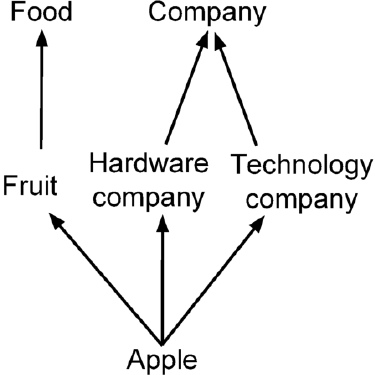}
\caption{An example DAG built using generalizations of term \textit{apple}. }
\label{fig:apple_hyp}
\end{figure}
\paragraph{\textbf{Model.}}We now describe our model for extracting hypernym subsequences for a given term.
We begin with a general formulation using directed acyclic graphs (referred to as DAG), and we make simplifying assumptions to derive a model for hypernym subsequences. 
We use the following notations:

\begin{itemize}[leftmargin=0.2cm,noitemsep,topsep=0pt]
\setlength{\itemindent}{0.8em}
\item $t_0$: a given seed term, e.g., \textit{apple};
\item $l_t$: lexical head of any term $t$, e.g., $l_t$=\textit{soup} for \mbox{$t$=\textit{chicken soup}};
\item $E$: Hypernym \underline{\textit{E}}vidence, i.e., the set of all the candidate hypernymy relations, in the form of 3-tuples (\textit{hyponym, hypernym, frequency});
\item $E_k(t)$: Hypernym \underline{\textit{E}}vidence for term $t$, i.e., the set of top-\underline{$k$} candidate hypernyms for term $t$, having the highest frequency counts (Table~\ref{tab:apple_hyp} shows a sample from $E_k(t)$ for $t$=\textit{apple});
\item $E_k(t, m)$: \textit{m}$^{th}$ ranked candidate hypernym from $E_k(t)$, where $m\leq k$, and ranks are computed by sorting candidate hypernyms in decreasing order of frequency counts;
\item $\text{sim}(t_i, t_j)$: A similarity measure between terms $t_i$ and $t_j$ estimated using evidence $E$;
\item $G_{t}$: a DAG consisting of generalizations for a term $t$ (Figure~\ref{fig:apple_hyp} shows an example of a possible DAG for $t$$=$\textit{apple}).
\end{itemize}

\vspace{0.3cm}
For a given term $t_0$, we define the goal of this step of our taxonomy induction approach as finding a DAG $\hat{G}_{t_0}$, which maximizes the conditional probability of $G_{t_0}$, given the evidence $E_k(t_0)$, for a fixed $k$:
\begin{eqnarray}
\hat{G}_{t_0}&=&\underset{G_{t_0}}{\text{argmax}}\;\text{Pr}(G_{t_0}|E_k({t_0})) \nonumber \\ 
&=&\underset{G_{t_0}}{\text{argmax}}\;\text{Pr}(E_k(t_0)|G_{t_0}) \times \text{Pr}(G_{t_0}) \label{eqn:1}
\end{eqnarray}
Due to the combinatorial nature of the search space of $G_{t_0}$, finding an exact solution to the above equation is intractable, even for a small $k$. Therefore, we make the following simplifying assumptions, which
facilitate an efficient search through the search space of $G_{t_0}$:
\begin{itemize}[leftmargin=0.2cm,itemsep=1pt,topsep=5pt]
\setlength{\itemindent}{0.8em}
\item$G_{t_0}$ can be approximated as a set of independent hypernym subsequences with possibly repeated hypernyms. In other words,  $G_{t_0}=\bigcup_{i=1}^{b} S_{t_0}^i$ where $S_{t_0}^i$ is the $i^{\text{th}}$ subsequence and $b$ is a fixed constant. For example, the DAG shown in Figure~\ref{fig:apple_hyp} can be approximated as a set of three subsequences: (i) \textit{apple}$\rightarrow$\textit{fruit}$\rightarrow$\textit{food}, (ii) \textit{apple}$\rightarrow$\textit{hardware company}$\rightarrow$\textit{company}, and (iii) \textit{apple}$\rightarrow$\textit{technology company}$\rightarrow$\textit{company}. This assumption intuitively derives from the fact that any DAG can be represented by a finite number of subsequences. 
\item$\forall i$, the joint events $(E_k(t_0), S_{t_0}^i)$ are independent. Intuitively, this assumption implies that each subsequence independently contributes to the evidence $E_k(t_0)$.
\item$\forall i$, the direct hypernyms of $t_0$ in $S_{t_0}^i$ are unique. In other words, for a candidate hypernym $h_c$ of given term $t_0$, there is at most one subsequence with the first edge $t_0$$\rightarrow$$h_c$. Intuitively, this assumption implies that a candidate hypernym $h_c$ uniquely sense-disambiguates the term $t_0$, thus resulting in a only one possible generalization subsequence.  
\end{itemize}

\vspace{0.25cm}
In conjunction, these assumptions imply that $G_{t_0}$ is composed of $b$ hypernym subsequences, where each subsequence independently attempts to generate $E_k({t_0})$. Given these assumptions, Equation~\ref{eqn:1} transforms into:
\begin{eqnarray}
\hat{G}_{t_0}&=&\underset{\bigcup_{i=1}^{b} S_{t_0}^i}{\text{argmax}}\;\prod_{i=1}^{b}\text{Pr}(E_k({t_0})|S_{t_0}^i)\times \text{Pr}(S_{t_0}^i)
\label{eqn:2}
\vspace{-2cm}
\end{eqnarray}

\vspace{0.2cm}
\paragraph{\textbf{Estimation.}}We now describe the estimation of $\text{Pr}(E_k({t_0})|S_{t_0}^i)$ and $\text{Pr}(S_{t_0}^i)$ for a hypernym subsequence ${S_{t_0}^i}$. In order to motivate the estimation of the conditional probability $\text{Pr}(E_k({t_0})|S_{t_0}^i)$, we start with an example. Consider a valid hypernym subsequence
\textit{apple}$\rightarrow$\textit{fruit}$\rightarrow$\textit{food}$\rightarrow$\textit{substance}$\rightarrow$\textit{matter}$\rightarrow$\textit{entity} for the term \textit{apple} (whose candidate hypernyms are in Table~\ref{tab:apple_hyp}). 
At first sight, it might seem desirable for a candidate hypernym from $E_k(t_0)$ (e.g., \textit{fruit}) to have a high similarity with as many terms in the subsequence as possible. 
However, since the similarity measure is based on the hypernym evidence $E$, it is plausible that terms such as \textit{matter} and \textit{entity} have a low similarity with the candidate hypernym \textit{fruit}, simply because they are at a higher level of generality.
To avoid penalizing such valid subsequences, we let the conditional probability $\text{Pr}(E_k({t_0})|S_{t_0}^i)$ be proportional to the maximum similarity possible between the candidate hypernym and \textit{any} term in the subsequence (e.g., for the candidate hypernym \textit{fruit}, the similarity is 1 as \textit{fruit} is in the subsequence). We aggregate those similarity values across the candidate hypernyms.
More formally, assuming subsequence \mbox{$S_{t_0}^i$ = $t_0$$\rightarrow$$h_{i1}$$\rightarrow$$h_{i2}$\dots$h_{in}$}, where $n$ is the length of $S_{t_0}^i$, we compute the conditional probability as:
\begin{eqnarray}
\text{Pr}(E_k({t_0})|S_{t_0}^i)\propto\sum_{m=1}^{k}({\lambda}_1)^m\underset{j\in \lbrack 1,n\rbrack}{\max}\big(\text{sim}(E_k(t_0,m),h_{ij})\big)\label{eqn:4}
\end{eqnarray}
where $\lambda_1$ (a fixed parameter) serves as a rank-penalty to penalize candidate hypernyms with lower frequency counts.\\
\newline
We now proceed to compute $\text{Pr}(S_{t_0}^i)$, the other constituent of Equation~\ref{eqn:2}. Towards that, we assume that $S_{t_0}^i$ is a collection of independent hypernym edges. Thus, $\text{Pr}(S_{t_0}^i)$ becomes the product of the individual edges' probabilities:
\begin{eqnarray}
\text{Pr}(S_{t_0}^i) \propto {\text{Pr}_{e}({t_0},h_{i1})\times (\lambda}_2)^n  \prod_{j=1}^{n-1}\text{Pr}_{e}(h_{ij},h_{i(j+1)})\label{eqn:5}
\end{eqnarray}
where $\text{Pr}_{e}(x_1,x_2)$ is the probability of an individual hypernym edge $x_1$$\rightarrow$$x_2$ between terms $x_1$ and $x_2$; ${\lambda}_2$ is a length penalty parameter.     
\newline
Finally, we estimate $\text{Pr}_e(x_1,x_2)$ as a log-linear model using a set of features \textbf{\mbox{f}}, weighted by the learned weight vector \textbf{w}:
\begin{eqnarray}
\text{Pr}_e(x_1,x_2) &\propto& \exp\big(\textbf{w} \cdot \textbf{f}(x_1, x_2)\big)\label{eqn:features}
\end{eqnarray}
We also use this edge probability to compute the aforementioned similarity function ($\text{sim}$) as:
\begin{eqnarray}
\text{sim}(x_i,x_j) &=& \max\big(\text{Pr}_e(x_i, x_j), \text{Pr}_e(x_j, x_i)\big)  \label{eqn:sim}
\end{eqnarray}
\newline
Intuitively, $\text{Pr}(E_k(t_0)|S_{t_0}^i)$ promotes subsequences containing a larger number of candidate hypernyms from $E_k({t_0})$ whereas $\text{Pr}(S_{t_0}^i)$ promotes subsequences consisting of individual edges with a larger probability of hypernymy.

\vspace{0.2cm}
\paragraph{\textbf{Subsequence Extraction.}}
After inserting Equations~\ref{eqn:4} and~\ref{eqn:5} into Equation~\ref{eqn:2} and taking logarithm, the objective function becomes:
\begin{eqnarray}
\begin{aligned}
&\hat{G}_{t_0}=\underset{\bigcup_{i=1}^{b} S_{t_0}^i}{\text{argmax}}\;\sum_{i=1}^{b} \Big[ \log \sum_{m=1}^{k}({\lambda}_1)^m \underset{j\in \lbrack 1,n\rbrack}{\max}\big(\text{sim}(E_k(t_0,m),h_{ij})\big) \\&+ \log\text{Pr}_e(t_0,h_{i1}) + n\lambda_2 + \sum_{j=1}^{n-1}\log\text{Pr}_e(h_{ij},h_{i(j+1)})\Big]
\nonumber\label{eqn:6}
\end{aligned}
\end{eqnarray}

This objective function leads to the following search algorithm for the extraction of subsequences: 
\begin{enumerate}
\item
For a given term $t_0$, iterate over all candidate hypernyms in $E_k(t_0)$.
\item
For each $h_c\in E_k(t_0)$, perform a depth-limited beam search over the space of possible subsequences by recursively exploring the candidate hypernyms of $h_c$ (i.e., $E_k(h_c))$.
\item
For each $h_c\in E_k(t_0)$, choose the subsequence $S$ with the highest score (i.e., $\log( \text{Pr}(E_k(t_0)|S)\times\text{Pr}(S)))$.
\item
Choose the top-$b$ candidate hypernyms based on their corresponding subsequence scores. 
\end{enumerate}

While, in theory, we can iterate over all candidate hypernyms in $E_k(t_0)$, in practice, we employ an alternative two-stage execution that significantly improves the running time as well as produces more meaningful subsequences:
\vspace{0.2cm}
\newline\forceindent$\bullet$ \textit{Search phase}: Proceed as in the aforementioned steps. However, in the special case where a candidate hypernym $h_c$ is a compound term and its lexical head $l_{h_c}$ is also present in $E_k(t_0)$, skip $h_c$ in step (1) of the algorithm\footnote{Lexical heads of terms have consistently played a special role in taxonomy induction~\citep{ponzetto2011taxonomy,guptarevisiting}.}. For example, for $t_0$ $=$ \textit{apple}, candidate hypernyms \textit{tech company}, \textit{software company} and \textit{hardware company} are skipped in step (1) due to the presence of \textit{company} in $E_k(t_0)$ (cf. Table~\ref{tab:apple_hyp}).
\vspace{0.2cm}
\newline\forceindent$\bullet$ \textit{Expansion phase}: In this phase, we augment the subsequences extracted in the search phase to account for skipped compound terms. We focus on the case where the lexical head of the skipped compound terms occurs in a subsequence. In that case, we expand the incoming edge of the lexical head with zero or more of those compound terms. For example, in the subsequence \textit{apple}$\rightarrow$\textit{company}$\rightarrow$\textit{organization}, a potential expansion of the edge \textit{apple}$\rightarrow$\textit{company} is: \textit{apple}$\rightarrow$\textit{American software company}$\rightarrow$\textit{software company}$\rightarrow$\textit{company}. 
However, special attention has to be taken while generating these potential expansions. For example, the expansion \textit{apple}$\rightarrow$\textit{American software company}$\rightarrow$\textit{British software company}$\rightarrow$\textit{company} is invalid due to the co-hyponymy edge \textit{American software company}$\rightarrow$\textit{British software company}. In contrast, the expansion {\textit{apple}$\rightarrow$\textit{American software company}$\rightarrow$\textit{software company}$\rightarrow\,$\textit{company} is a valid expansion. 
To avoid invalid expansions, we restrict the possible expansions to the case where the set of pre-modifiers of a compound term is a superset of its hypernym's pre-modifiers (e.g., \mbox{\{\textit{American, software} \}$\supset$\{\textit{software}\}}).

We generate all possible expansions for each edge and rank them by averaging a TF-IDF-style metric across the pre-modifiers of compound terms in each expansion. 
Our aim in the ranking is two-fold: i) promoting the pre-modifiers, which frequently appear in the evidence $E_k(t_0)$, and ii) penalizing the noisy pre-modifiers unrelated to $t_0$ that frequently occur in compound terms (e.g., \textit{several}, \textit{other}, etc.). Hence, we compute the TF score of a pre-modifier as its average frequency of occurrence in the candidate hypernyms $E_k(t_0)$. We compute IDF as the average frequency of occurrences of the pre-modifier in $E_k(t)$ for a random term $t$. Finally, we choose the top ranked expansion per edge.

To illustrate the result of the previous steps, we show in Table~\ref{tab:subseq} an example of extracted subsequences along with their expanded versions for the food domain. Intuitively, the two-stage execution serves to distinguish between two fundamentally different forms of generalization:
\begin{enumerate}[leftmargin=0.8cm,itemsep=3pt,topsep=1pt]
\item
\textbf{type-based generalization}, which provides core types as generalizations (e.g.,  \textit{apple}$\rightarrow$\textit{company}$\rightarrow$\textit{organization}).
\item
\textbf{attribute-based generalization}, which enriches type-based generalization edges. For example, \textit{apple}$\rightarrow$\textit{american software company}$\rightarrow$\textit{software company}$\rightarrow$\textit{company} enriches the individual type-based edge \textit{apple}$\rightarrow$\textit{company}.
\end{enumerate}

In our experiments, models that distinguished between these two different forms of generalizations consistently performed better than models, which attempted to unify them. 

\paragraph{\textbf{Features.}} We now describe the edge features that we employ for estimating the probability of a hypernymy relation between two terms (cf. Equation~\ref{eqn:features}):
\newline\forceindent$\bullet$ \textit{Normalized Frequency Diff ($n_d$)}: Similar to~\cite{panchenko2016taxi}, this feature is an asymmetric hypernymy score based on frequency counts.
We compute $n_d(x_i, x_j)$ by first normalizing the frequency counts obtained (i.e., the counts in $E_k(x_i)$) for  term $x_i$ as follows: $n_f(x_i, x_j) = \frac{\text{freq}(x_i,x_j)}{\underset{m}{\text{max}}\; \text{freq}(x_i,x_m)}$, where $\text{freq}(x_i,x_j)$ is the frequency count of candidate hypernym $x_j$ in $E_k(x_i)$. Further, we subtract the score in the opposite direction to downrank synonyms and co-hyponyms: $n_d(x_i, x_j) = n_f(x_i, x_j) - n_f(x_j, x_i)$.
\newline\forceindent$\bullet$ \textit{Generality Diff ($g_d$)}: We introduce a novel feature for explicitly incorporating the term generality (or abstractness) in our model. To this end, we first define the generality $g(t)$ of a term $t$ as the log of the number of distinct hyponyms present in all candidate hypernymy relations ($E$); i.e., $g(t) = \text{log}(1 + \lvert x \mid x$$\rightarrow$$t \in E \rvert)$. 
We define the generality of an edge as the difference in generality between the hypernym and the hyponym: $g_e(x_i, x_j) = g(x_j) - g(x_i)$. 
\begin{table}
\resizebox{0.5\textwidth}{!}{
\begin{tabular}{l}
    \toprule
    \textbf{Initial subsequences}  \\
    \midrule
    \textit{mortadella}$\rightarrow$\textit{sausage}$\rightarrow$\textit{meat}$\rightarrow$\textit{food}\\
   \textit{laksa}$\rightarrow$\textit{soup}$\rightarrow$\textit{dish}$\rightarrow$\textit{food}\\
   \toprule
   \textbf{Expanded subsequences} \\
    \midrule
\textit{mortadella}$\rightarrow$\textit{large Italian sausage}$\rightarrow$\textit{sausage}$\rightarrow$\textit{process meat}$\rightarrow$\textit{meat}$\rightarrow$\textit{food}\\
 \textit{laksa}$\rightarrow$\textit{spicy noodle soup}$\rightarrow$\textit{noodle soup}$\rightarrow$\textit{soup}$\rightarrow$\textit{dish}$\rightarrow$\textit{food}\\
  \end{tabular}
  }
 \caption{Examples of hypernym subsequences found during the search phase, and their expanded versions.}
\label{tab:subseq}
  \vspace{-0.3cm}
  \end{table}

  \input{figures/taxonomy-fig}

Intuitively, we aim to promote edges with the right level of generality and penalize edges, which are either too general (e.g., \mbox{\textit{apple}$\rightarrow$\textit{thing}}) or too specific (i.e., edges between synonyms or co-hyponyms, such as \mbox{\textit{apple}$\rightarrow$\textit{orange}}). 
To realize this intuition, we first sample a random set of terms and collect the edges with highest $n_d$ for these terms (hereafter referred to as \textit{top edges}). We compare the distribution of generality (i.e., $g_e$) for the top edges vs. the distribution of generality for a set of randomly sampled edges. 
The assumption is that it is more likely to sample the generality of a correct edge (i.e., edge at right level of generality) from the distribution of top edges as compared to random edges. Hence, given $D_t$ and $D_r$ as the Gaussian distributions estimated from the samples of generality for top edges and random edges respectively, we define the feature as: $g_d(x_i, x_j) = \text{Pr}_{D_t}\big(g_e(x_i, x_j)\big) - \text{Pr}_{D_r}\big(g_e(x_i, x_j)\big)$. 

\paragraph{\textbf{Parameter Tuning.}}We estimate the weights for features (\textbf{w} in equation~\ref{eqn:features}), using a support vector machine trained on a manually annotated set of 500 edges. For beam search in the search phase, we use a beam of width 20, and limit the search to subsequences of maximum length 4.  We set the rest of the parameters by running grid-search over a manually-defined range of parameters using a small validation set\footnote{Validation set is excluded from the test set.}. The final values of parameters are as follows: $k$$=$$10$, $b$$=$$4$, $\lambda_{1}$$=$$\lambda_{2}$$=$$0.95$. 

\subsection{Aggregation of Subsequences}
\label{sec:agg}
Up till now, we have described our methodology to generate hypernym subsequences starting from a given term. In this section, we aggregate the hypernym subsequences obtained for a set of seed terms, in order to construct an initial hypernym graph.
For that, we undertake the following steps:

\paragraph{\textbf{Domain Filtering.}} Given a term $t_0$, the usual case is that multiple hypernym subsequences corresponding to different senses of the term $t_0$ are extracted. For example, \textit{apple} can be a \textit{company} or a \textit{fruit}, thus resulting in subsequences \textit{apple}$\rightarrow$\textit{fruit}$\rightarrow$\textit{food} and \textit{apple}$\rightarrow$\textit{software company}$\rightarrow$\textit{company}. However, many of these subsequences will not pertain to the domain of interest (as determined by the seed terms). To eliminate the irrelevant ones, we estimate a smoothed unigram model\footnote{We used a weighting function (i.e., step function with cut-off at 50\% of the height of the subsequence) to favor terms at lower heights as they are usually more domain-specific.} from all extracted subsequences, and we remove those with generation probabilities below a fixed threshold. 
\paragraph{\textbf{Hypernym Graph Construction.}} 
We now aggregate the filtered subsequences into an initial hypernym graph. We construct this graph by grouping the edges with the same start and end nodes together from the filtered subsequences. The weight of each edge is computed as the sum of the scores of subsequences it belongs to (i.e., $\log$( $\text{Pr}(E_k(t)|S)\times \text{Pr} (S))$). To increase the coverage for compound seed terms that do not yet have a hypernym, we simply add an hypernym edge to their lexical head with weight$=$$\infty$ (i.e, a very large value) whenever the lexical head is already present in the hypernym graph. Finally, for each cycle in the hypernym graph, we remove the edge with the smallest weight, hence resulting in a DAG.  This DAG contains many noisy terms and edges, which are pruned in the next step of our approach.

\subsection{Taxonomy Construction}

\label{sec:flow}

In this step, we aim to induce a tree-like taxonomy from the hypernym DAG obtained in the previous step. We cast this as an instance of the minimum-cost flow  problem (MCFP).

MCFP is an optimization problem, which aims to find the cheapest way of sending a certain amount of flow through a flow network. It has been used to find the optimal solution in applications like the \textit{transportation problem}~\citep{klein1967primal}, where the goal is to find the cheapest paths to send commodities from a group of facilities to the customers via a transportation network. Analogously, we cast the problem of taxonomy induction as finding the cheapest way of sending the seed terms to the root terms through a carefully designed flow network $F$. 
We use the \textit{network simplex algorithm}~\citep{orlin1997polynomial} to compute the optimal flow for $F$, and we select all edges with positive flow as part of our final taxonomy.
We now describe our method for constructing the flow network $F$. In what follows, we refer to Figure~\ref{fig:taxonomy-induction} at the different steps.

\paragraph{\textbf{Flow Network Construction.}} Let $V$ be the vocabulary of input seed terms (e.g., \textit{apple}, \textit{orange}, and \textit{Spain} in Figure~\ref{fig:taxonomy-induction}); $H$ is the noisy hypernym graph constructed in Section ~\ref{sec:agg} (cf. Figure~\ref{fig:taxonomy-induction}(a)); $w(x, y)$ is the weight of the edge $x$$\rightarrow$$y$ in $H$; $D_x$ is the set of descendants of term $x$ in $H$ (e.g., \textit{apple} is a descendant of \textit{food}); $R$ is the set of given roots\footnote{If roots are not provided, a small set of upper terms can be used as roots~\citep{velardi2013ontolearn}.} (e.g., \textit{food} in Figure~\ref{fig:taxonomy-induction}).
The construction of the flow network $F$ proceeds as follows (cf. Figure~\ref{fig:taxonomy-induction}(b)):
\begin{enumerate}[label=\roman*),leftmargin=0.2cm,noitemsep,topsep=0pt]
\setlength{\itemindent}{1.4em}
\item For an edge $x$$\rightarrow$$y$ in $H$, add the edge $x$$\rightarrow$$y$ in $F$. Set the capacity ($c$) of the added edge as \mbox{$c(x,y)=\lvert D_x\cap V\rvert$}. 
Set the cost ($a$) of that edge as \mbox{$a(x,y)=1 / w(x,y)$}. 
\item Add a sentinel \textit{source} node $s$. $\forall v \in V$, add an edge $s$$\rightarrow$$v$ with $c(s,v)=a(s,v)=1$.

\item Add a sentinel \textit{sink} node $t$. $\forall r \in R$, add edge $r$$\rightarrow$$t$ with $c(r,t)=\lvert D_r\cap V\rvert$ and $a(r,t)=1$.
\end{enumerate}

\paragraph{\textbf{Minimum-cost Flow.}} Given a demand $d$ of the total flow to be sent from $s$ to $t$, the goal of MCFP is to find flow values ($f$) for each edge in $F$ that minimize the total cost of flow over all edges: $\underset{(u,v) \in F}{\sum}a(u,v)\cdot f(u,v)$. In our construct, demand $d$ represents the maximum number of seed terms that can be included in the final taxonomy. Figures~\ref{fig:taxonomy-induction}(c) and~\ref{fig:taxonomy-induction}(d) show the minimum-cost flow for demand $d$$=$3 and $d$$=$2 respectively. In both cases, the edge \textit{apple}$\rightarrow$\textit{food} receives $f$$=$0 due to the presence of edges \textit{apple}$\rightarrow$\textit{fruit} and \textit{fruit}$\rightarrow$\textit{food} with lower costs. For $d$$=$2, the edge \textit{source}$\rightarrow$\textit{Spain} has $f$$=$0, implying that the noisy term \textit{Spain} would be removed from the final taxonomy. Intuitively, demand $d$ serves as a parameter for discarding potentially noisy terms in the input vocabulary. More formally, $d$ can be defined as $\alpha$$\lvert V\rvert$, where $\alpha$, a user-defined parameter, indicates the desired \textit{coverage} over seed terms. If the vocabulary contains only accurate terms, $\alpha$ is set to 1. 
For a given $\alpha$, we run the network simplex algorithm with $d$$=$$\alpha$$\lvert V\rvert$ to compute the minimum-cost flow for $F$. The final taxonomy consists of all edges with flow $>0$.

%% file: figures/taxonomy-fig.tex
\begin{figure*}[t]
\resizebox{\linewidth}{!}{
\begin{tikzpicture}[->,>=stealth',shorten >=1pt,auto]
  \tikzset{Category/.style ={shape=circle,fill=white,draw=black,minimum size=6mm,inner sep=0pt}}
  \tikzset{Discarded/.style ={shape=circle,fill=gray!50,draw=black,minimum size=6mm,inner sep=0pt}}
  \tikzset{Discarded2/.style ={shape=circle,pattern=crosshatch dots,pattern color=gray!30,draw=black,minimum size=6mm,inner sep=0pt}}
  \tikzset{Entity/.style ={shape=circle,fill=black,draw=black,text=white,minimum size=6mm,inner sep=0pt}}
  \tikzset{Name/.style = {text=black, node distance=4mm, align=center, scale=0.7, font=\relsize{3.7}}}
  \newcommand\NameW[2]{%
	\node[Name, left of=#1, anchor=east] (#1_t) {#2};
  }
  \newcommand\NameE[2]{%
	\node[Name, right of=#1, anchor=west] (#1_t) {#2};
  }
  \newcommand\NameNE[2]{%
	\node[Name, right of=#1, anchor=south west] (#1_t) {#2};
  }
  \newcommand\NameS[2]{%
	\node[Name, below of=#1, anchor=north, node distance=4mm] (#1_t) {#2};
  }
  \newcommand\NameN[2]{%
	\node[Name, above of=#1, anchor=south, node distance=4mm] (#1_t) {#2};
  }
  \newcommand{\EdgeSolid}[2]{%
  \draw [->] (#1) -- (#2);
  }
  \newcommand{\EdgeStyle}[3]{%
    \path[#3] (#1) edge (#2);
  }
  \newcommand{\EdgeDotted}[2]{%
  \draw [dotted,->] (#1) -- (#2);
  }

\begin{scope}[shift={(-3.8,-1.0)}]
   \node[text width=6cm] (group_a)	at (1.5, -2) {\LARGE(a): Noisy hypernym graph (H).};
   \node[Discarded]    (apple)	at (-0.5, -0.2) {};
   \node[Discarded]  (orange)		at (2.5, -0.2) {};
   \node[Category]  (fruit)	at (1, 1.2) {};
   \node[Discarded2]  (food)	at (1, 2.6) {};
   \node[Discarded]  (spain)	at (3, 1) {};
   
    \NameS{apple}{\\ apple}
    \NameW{orange}{\\orange}
    \NameE{fruit}{\\ fruit}
    \NameW{food}{\\ food}
    \NameE{spain}{\\ Spain}
       
       \begin{scope}[every path/.style={->}, every node/.style={sloped, inner sep=1pt}]
 \draw (apple) -- node [anchor=north] {{\Large 0.5}} (fruit);
 \draw (apple) -- node [anchor=south, pos=0.5] {{\Large 0.2}} (food);
 \draw (orange) -- node [anchor=north] {{\Large $0.5$}} (fruit);
 \draw (spain) -- node [anchor=south] {{\Large $0.25$}} (food);
 \draw (fruit) -- node [anchor=north] {{\Large $1$}} (food);
 \end{scope}
  \end{scope}

 \begin{scope}[shift={(3,-1.0)}]
   \node[text width=7cm] (group_b)	at (1.5, -2) {\LARGE(b): Flow network $F$ with (capacity, cost) values for each edge.};
   \node[Entity]    (start)	at (3.6, -1) {};
   \node[Entity]    (end)	at (2.5, 2.7) {};
   
    \node[Discarded]    (apple)	at (-0.5, -0.2) {};
   \node[Discarded]  (orange)		at (2.5, -0.2) {};
   \node[Category]  (fruit)	at (1, 1.2) {};
   \node[Discarded2]  (food)	at (1, 2.6) {};
   \node[Discarded]  (spain)	at (3, 1) {};
   
    \NameNE{start}{\\ source}
    \NameE{end}{\\ sink}
    \NameS{apple}{\\ apple}
    \NameW{orange}{\\orange}
    \NameE{fruit}{\\ fruit}
    \NameW{food}{\\ food}
      \NameE{spain}{\\ Spain} 
       \begin{scope}[every path/.style={->}, every node/.style={sloped, inner sep=1pt}]
 \draw (apple) -- node [anchor=north] {{\Large (1, 2)}} (fruit);
 \draw (apple) -- node [anchor=south, pos=0.5] {{\Large (1, 5)}} (food);
 \draw (orange) -- node [anchor=north] {{\Large (1, 2)}} (fruit);
 \draw (spain) -- node [anchor=south] {{\Large (1, 4)}} (food);
 \draw (fruit) -- node [anchor=north] {{\Large (2, 1)}} (food);
 \end{scope}
  
 \begin{scope}[every path/.style={->, dashed}, every node/.style={sloped, inner sep=1pt}]
 \draw (start) -- node [anchor=north] {{\Large (1, 1)}} (apple);
 \draw (start) -- node [anchor=south] {{\Large (1, 1)}} (orange);
 \draw (start) -- node [anchor=south] {{\Large (1, 1)}} (spain);
 \draw (food) -- node [anchor=south] {{\Large (3, 1)}} (end);
  \end{scope}

 \end{scope}

 \begin{scope}[shift={(9.8,-1.0)}]
   \node[text width=6cm] (group_c)	at (1.5, -2) {\LARGE(c): Flow values ($f$) for each  edge found using demand $d=3$.};
   \node[Entity]    (start)	at (3.6, -1) {};
   \node[Entity]    (end)	at (2.5, 2.7) {};
   
    \node[Discarded]    (apple)	at (-0.5, -0.2) {};
   \node[Discarded]  (orange)		at (2.5, -0.2) {};
   \node[Category]  (fruit)	at (1, 1.2) {};
   \node[Discarded2]  (food)	at (1, 2.6) {};
   \node[Discarded]  (spain)	at (3, 1) {};
   
    \NameNE{start}{\\ source}
    \NameE{end}{\\ sink}
    \NameS{apple}{\\ apple}
    \NameW{orange}{\\orange}
    \NameE{fruit}{\\ fruit}
    \NameW{food}{\\ food}
    \NameE{spain}{\\ Spain}
       
       \begin{scope}[every path/.style={->}, every node/.style={sloped, inner sep=1pt}]
 \draw (apple) -- node [anchor=north] {\Large 1} (fruit);
 \draw (orange) -- node [anchor=north] {\Large 1} (fruit);
 \draw (spain) -- node [anchor=south] {\Large 1} (food);
 \draw (fruit) -- node [anchor=north] {\Large 2} (food);
 \draw (apple) -- node [anchor=south] {\Large 0} (food);
 \end{scope}
 
 \begin{scope}[every path/.style={->, dashed}, every node/.style={sloped, inner sep=1pt}]
 \draw (start) -- node [anchor=north] {\Large 1} (apple);
 \draw (start) -- node [anchor=south] {\Large 1} (orange);
 \draw (start) -- node [anchor=south] {\Large 1} (spain);
 \draw (food) -- node [anchor=south] {\Large 3} (end);
  \end{scope}
    \end{scope}

   \begin{scope}[shift={(17,-1.0)}]
   \node[text width=6cm] (group_d)	at (1.5, -2) {\LARGE (d): Flow values ($f$) for each  edge found using demand $d=2$.};
   \node[Entity]    (start)	at (3.6, -1) {};
  \node[Entity]    (end)	at (2.5, 2.7) {};
  
  \node[Discarded]    (apple)	at (-0.5, -0.2) {};
   \node[Discarded]  (orange)		at (2.5, -0.2) {};
   \node[Category]  (fruit)	at (1, 1.2) {};
   \node[Discarded2]  (food)	at (1, 2.6) {};
   \node[Discarded]  (spain)	at (3, 1) {};
   
    \NameNE{start}{\\ source}
    \NameE{end}{\\ sink}
    \NameS{apple}{\\ apple}
    \NameW{orange}{\\orange}
    \NameE{fruit}{\\ fruit}
    \NameW{food}{\\ food}
    \NameE{spain}{\\ Spain}   
       \begin{scope}[every path/.style={->}, every node/.style={sloped, inner sep=1pt}]
 \draw (apple) -- node [anchor=north] {\Large 1} (fruit);
 \draw (orange) -- node [anchor=north] {\Large 1} (fruit);
 \draw (fruit) -- node [anchor=north] {\Large 2} (food);
  \draw (apple) -- node [anchor=south] {\Large 0} (food);
  \draw (spain) -- node [anchor=south] {\Large 0} (food);
 \end{scope}
 
 \begin{scope}[every path/.style={->, dashed}, every node/.style={sloped, inner sep=1pt}]
 \draw (start) -- node [anchor=north] {\Large 1} (apple);
 \draw (start) -- node [anchor=south] {\Large 1} (orange);
 \draw (food) -- node [anchor=south] {\Large 2} (end);
 \draw (start) -- node [anchor=south] {\Large 0} (spain);
  \end{scope}
    \end{scope}
\end{tikzpicture}
}
  \caption{Execution of the minimum cost flow algorithm starting from our hypernym graph.}
  \label{fig:taxonomy-induction}
 
\end{figure*}
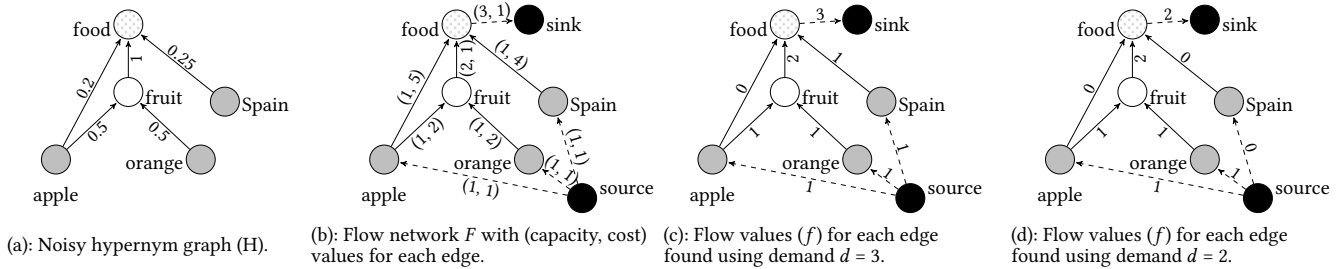

%% file: semeval.tex
\section{Evaluation}
\label{sec:eval}

The aim of the empirical evaluation is to address the following questions:
\begin{itemize}[leftmargin=0.35cm,noitemsep,topsep=1pt]
\item 
How does our approach compare to the state-of-the-art approaches under the assumption of a clean input vocabulary? 
\item
How does our approach perform on a noisy input vocabulary?
\item What are the benefits of extracting longer hypernym subsequences compared to single hypernym edges?
\end{itemize}
\vspace{0.25cm}
To this end, we perform two experiments. In Section~\ref{sec:sota}, we compare our taxonomy induction approach against the state of the art, under the simplifying assumption of a clean input vocabulary. Evaluations are performed automatically by computing standard precision, recall and F1 measures against a gold standard.

We then drop the simplifying assumption in Section~\ref{sec:noisy}, where we show that our taxonomy induction performs well even under the presence of significant noise in the input vocabulary. Evaluation is performed both manually as well as automatically against WordNet as the gold standard. We also demonstrate that the subsequences-based approach significantly outperforms an edges-based variant, thus demonstrating the utility of hypernym subsequences.

In the remainder of this section, we use \textit{SubSeq} to refer to our approach towards taxonomy induction (cf. Section~\ref{sec:tax}).

\subsection{Evaluation against the State of the Art}
\label{sec:sota}

\paragraph{\textbf{Setup.}}We use the setting of the SemEval 2016 task for taxonomy extraction~\citep{task13semeval2016}. The task provides 6 sets of input terminologies, related to three domains (food, environment and science),
for four different languages (English, Dutch, French and Italian).
The task requires participants to generate taxonomies for each (terminology, language) pair, which are further evaluated using a variety of techniques, including comparison against a gold standard. Except for a few restricted resources used to construct gold standard, the participants are allowed to use external corpora for hypernymy extraction and taxonomy induction. Participants are compared against each other and against a high-precision string inclusion baseline.

We compare SubSeq with TAXI, the system that reached the first place in all subtasks of the SemEval task~\citep{panchenko2016taxi}. TAXI harvests candidate hypernyms using substring inclusion and lexico-syntactic patterns from text corpora. It further utilizes an SVM trained with individual hypernymy edge features, such as frequency counts and substring inclusion to classify edges as positive and negative. The positive edges are added to the taxonomy. ~\citet{panchenko2016taxi} also report that alternate configurations of TAXI with different term-level and edge-level features as well as different classifiers such as Logistic Regression, Gradient Boosted Trees, and Random
Forest fail to provide improvements over their approach.   

In contrast to SubSeq, which discovers new hypernyms for the seed terms, SemEval task provides the additional assumption that all the terms in the gold standard taxonomies (i.e., including leaf terms and non-leaf terms) are present in the input vocabulary. This would unfairly lower the performance of SubSeq, as SubSeq would find hypernyms, which are possibly correct but not present in the gold standard. Hence, to ensure a fair comparison, we restrict the subsequence extraction and hypernym graph construction step of SubSeq (cf. Section~\ref{sec:tax}) to candidate hypernyms present in the input vocabulary. Furthermore, since candidate hypernymy extraction is orthogonal to our work, we reuse the candidate hypernymy relations made available by TAXI. As a consequence, TAXI and SubSeq are identical in input data conditions as well as evaluation metrics, and only differ in the core taxonomy induction approach.

\paragraph{\textbf{Evaluation Results.}}
\begin{table}[bt]
 \begin{tabular}
  {>{\scshape}r*{9}{>{\centering\arraybackslash}p{1.53em}}}
    \toprule
    &
\multicolumn{3}{c}{\textsc{TAXI}} & \multicolumn{3}{c}{\textsc{SubSeq}} \\
\cmidrule(lr){2-4}
    \cmidrule(lr){5-7}
  
    & \textsc{P} & \textsc{R} & \textsc{F1} & \textsc{P} & \textsc{R}  & \textsc{F1} \\
\midrule
EN & 33.2 & 31.7 & 32.2 & \textbf{44.9} & \textbf{31.9} & \textbf{37.2}\\
NL & \textbf{48.0} & 19.7 & 27.6 & 42.3 & \textbf{20.7} & \textbf{27.9}\\
FR & 33.4 & 24.1 & 27.7 & \textbf{41.0} & \textbf{24.4 } & \textbf{30.5}\\
IT & \textbf{53.7 }& 20.7 & 29.1 & 49.0 & \textbf{21.8}& \textbf{29.9}\\
    \bottomrule
    \end{tabular}
    \captionof{table}{Precision (P), Recall (R) and F1 Metrics for TAXI vs. SubSeq across different languages. Results are aggregated over all domains per language.}
    \label{tab:lang_com}
    \end{table}
    
\begin{figure}[tb]
    \centering
       \includegraphics[width=0.8\linewidth]{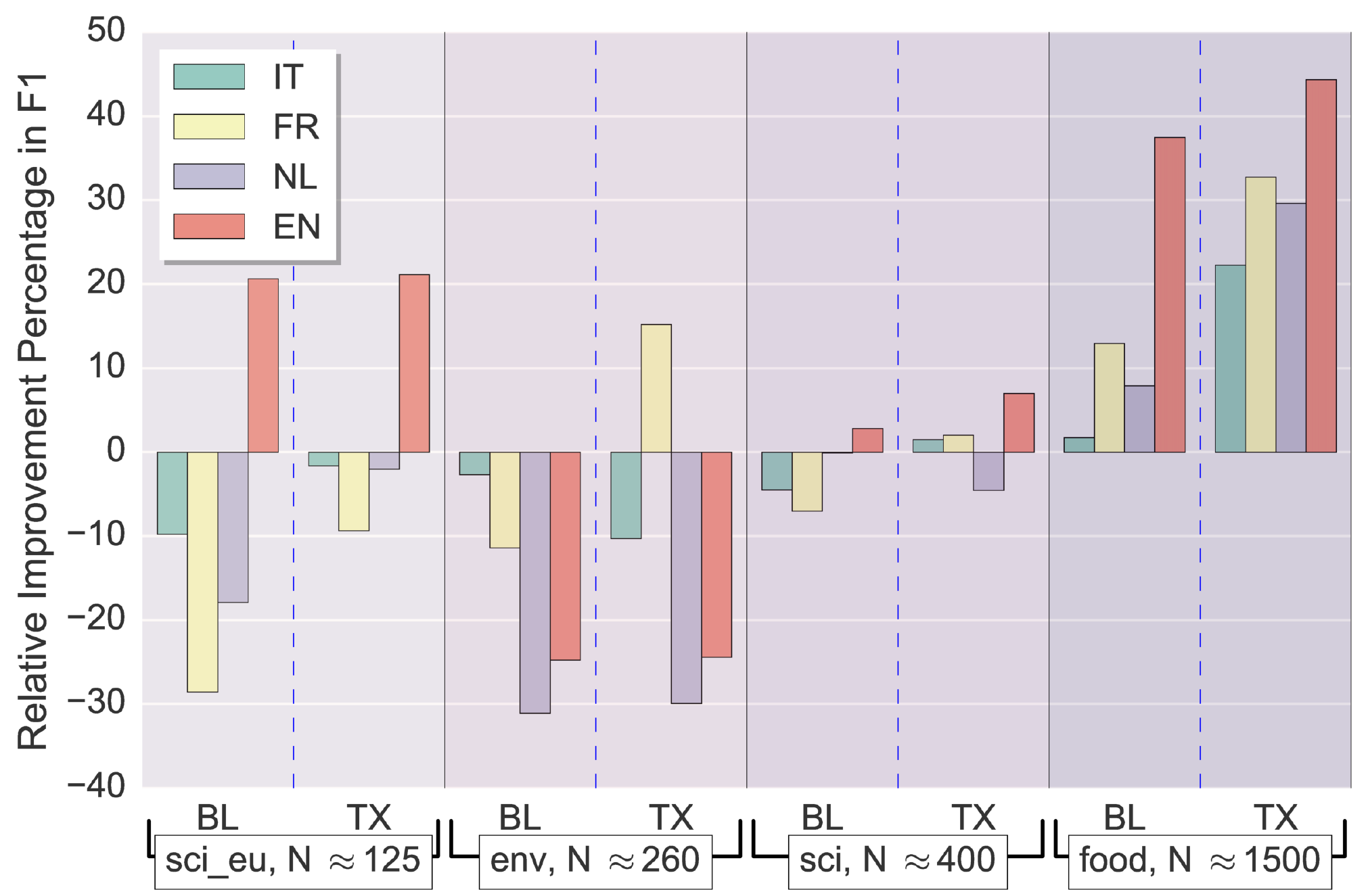}
       \caption{Relative improvement \% in F1 for SubSeq, compared to TAXI (TX) and the SemEval Baseline (BL), for different domains and languages. $N$ is the average number of terms in the input vocabulary for that domain. \textit{Science eurovoc} datasets are shown separately, as they have significantly fewer input terms than other science datasets. }
        \label{fig:grouped_barplot}
\end{figure}

Table~\ref{tab:lang_com} shows the language-wise precision, recall and F1 values computed against the gold standard for SubSeq and TAXI. Aggregated over all domains, SubSeq outperforms TAXI for all four languages. It achieves $>$15\% relative improvement in F1 for English and 7\% improvement overall. Both methods perform significantly better for English, which can be attributed to the higher accuracy of candidate hypernymy relations for English. Figure~\ref{fig:grouped_barplot} shows the performance of SubSeq compared to TAXI and the SemEval baseline across different domains and languages. SubSeq performs best for food domain, where it outperforms TAXI across all the languages. SubSeq performs best for English, where it outperforms TAXI across 3/4 domains. 

In our experiments, we noticed that SubSeq achieves the largest improvements when a greater number of hypernym subsequences are found during the subsequence extraction step. For example, SubSeq achieves an average 32.23\% relative improvement in F1 over TAXI for the food domain, where on an average 0.67 subsequences are found per term, compared to only 0.44 for the other domains. Similarly, SubSeq performs best for English datasets, where, on an average, 1.09 subsequences are found per term, compared to only 0.32 for other languages. The variation in the number of extracted subsequences per term can be attributed to two factors: (i) number of terms in the input vocabulary, and (ii) number of candidate hypernymy relations available. Due to the assumption that all candidate hypernyms belong to the input vocabulary, larger vocabularies of food domain make it more likely for a candidate hypernym to be found, and hence for a subsequence to be extracted. In a similar fashion, the larger set of available candidate hypernyms for English ($\sim$65 million vs. $<$ 2.2 million for other languages) makes it more likely for a subsequence to be extracted for English datasets.

Overall this experiment shows that under the assumption of a clean input vocabulary, SubSeq is more effective that TAXI for most domains in English, and domains with large vocabularies such as food in other languages.

%% file: realeval.tex

\begin{figure*}[tb]
    \centering
     
   \begin{minipage}{0.27\textwidth}
        \centering
        \includegraphics[width=0.9\linewidth]{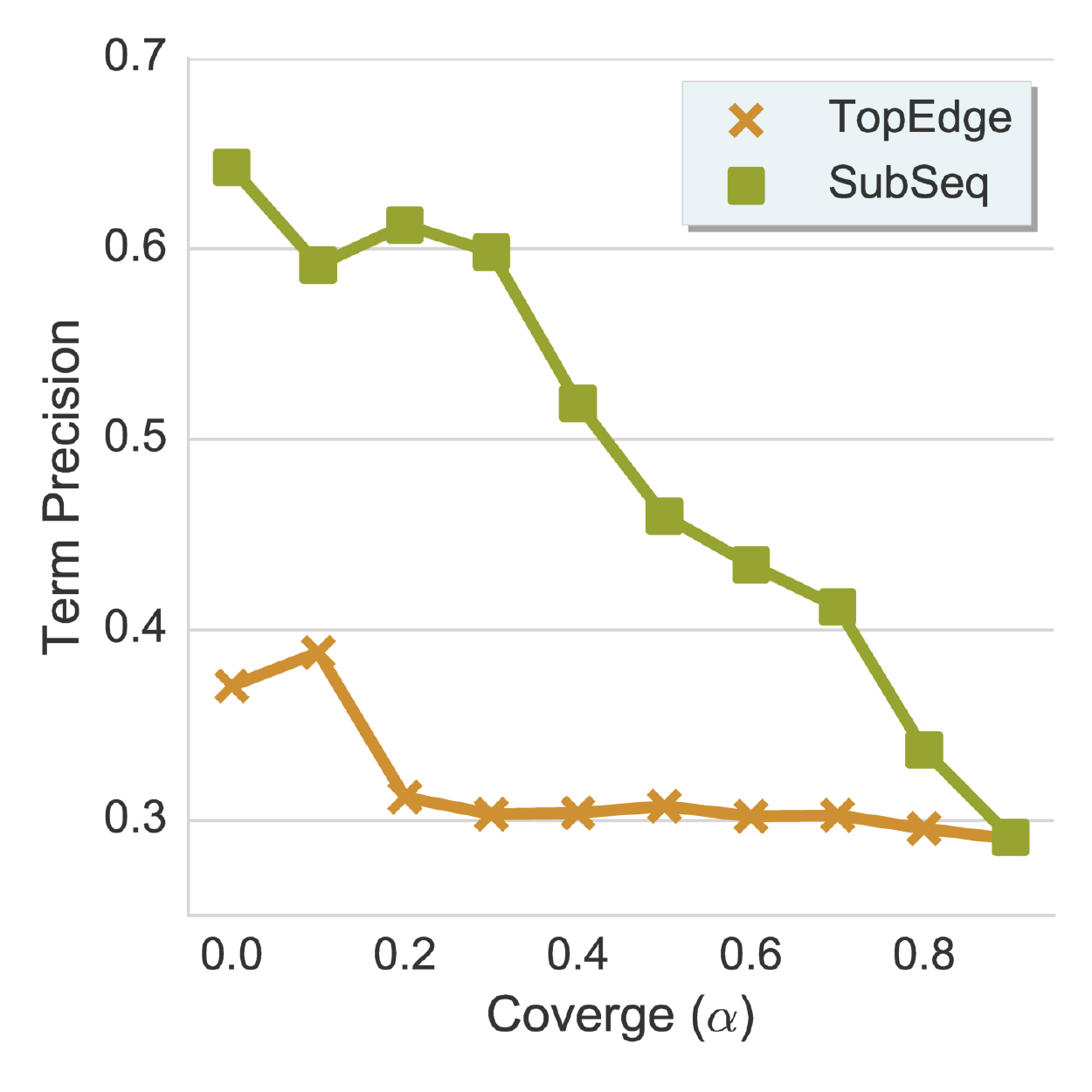}
        \caption{Term precision for SubSeq vs. TopEdge.}
        \label{fig:alpha3}
    \end{minipage}%
    \quad
    \begin{minipage}{.27\textwidth}
        \centering        \includegraphics[width=0.9\linewidth]{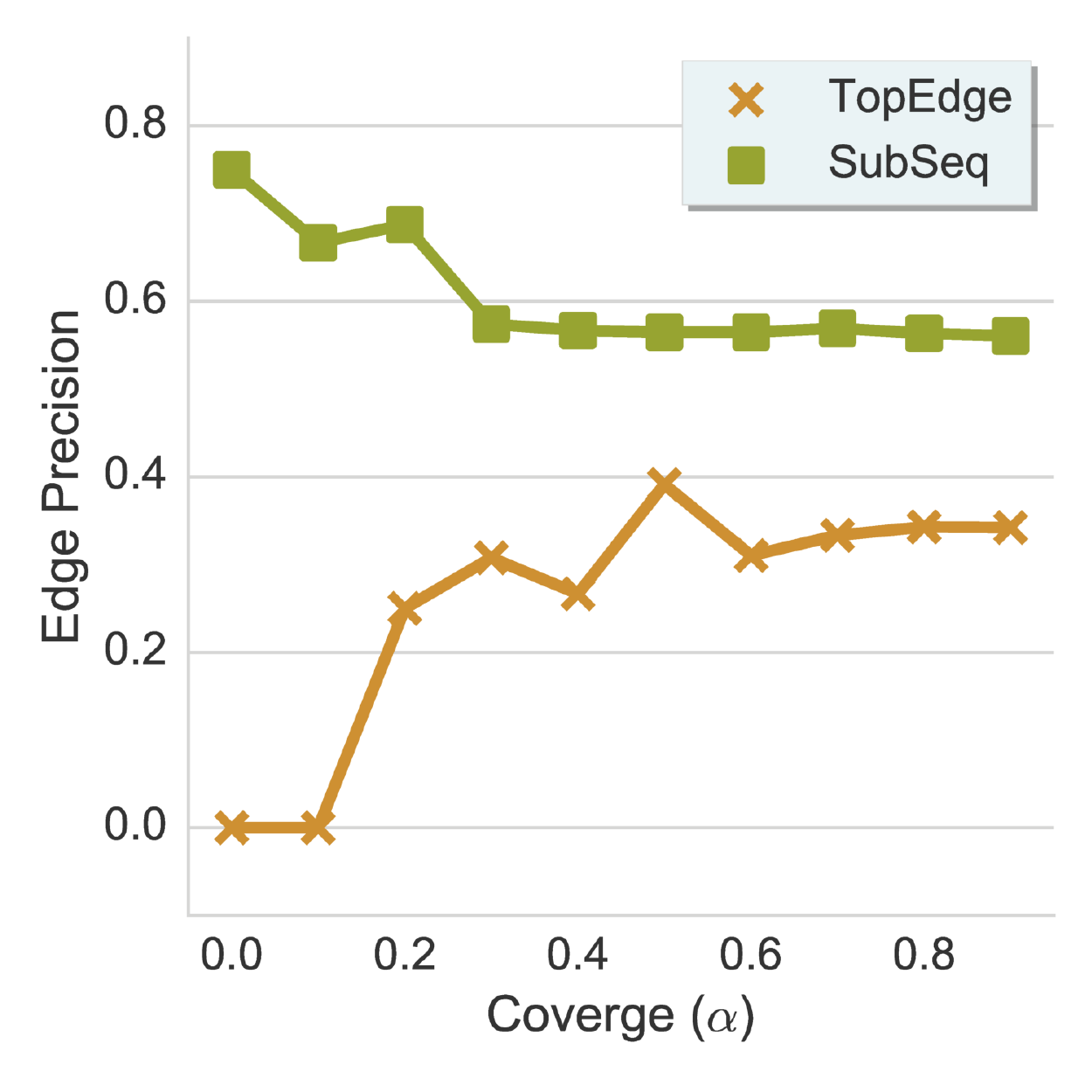}
        \caption{Edge precision for SubSeq vs. TopEdge.}
        \label{fig:alpha2}
    \end{minipage}%
    \quad
      \begin{minipage}{.4\textwidth}
        \centering        \includegraphics[width=0.9\linewidth]{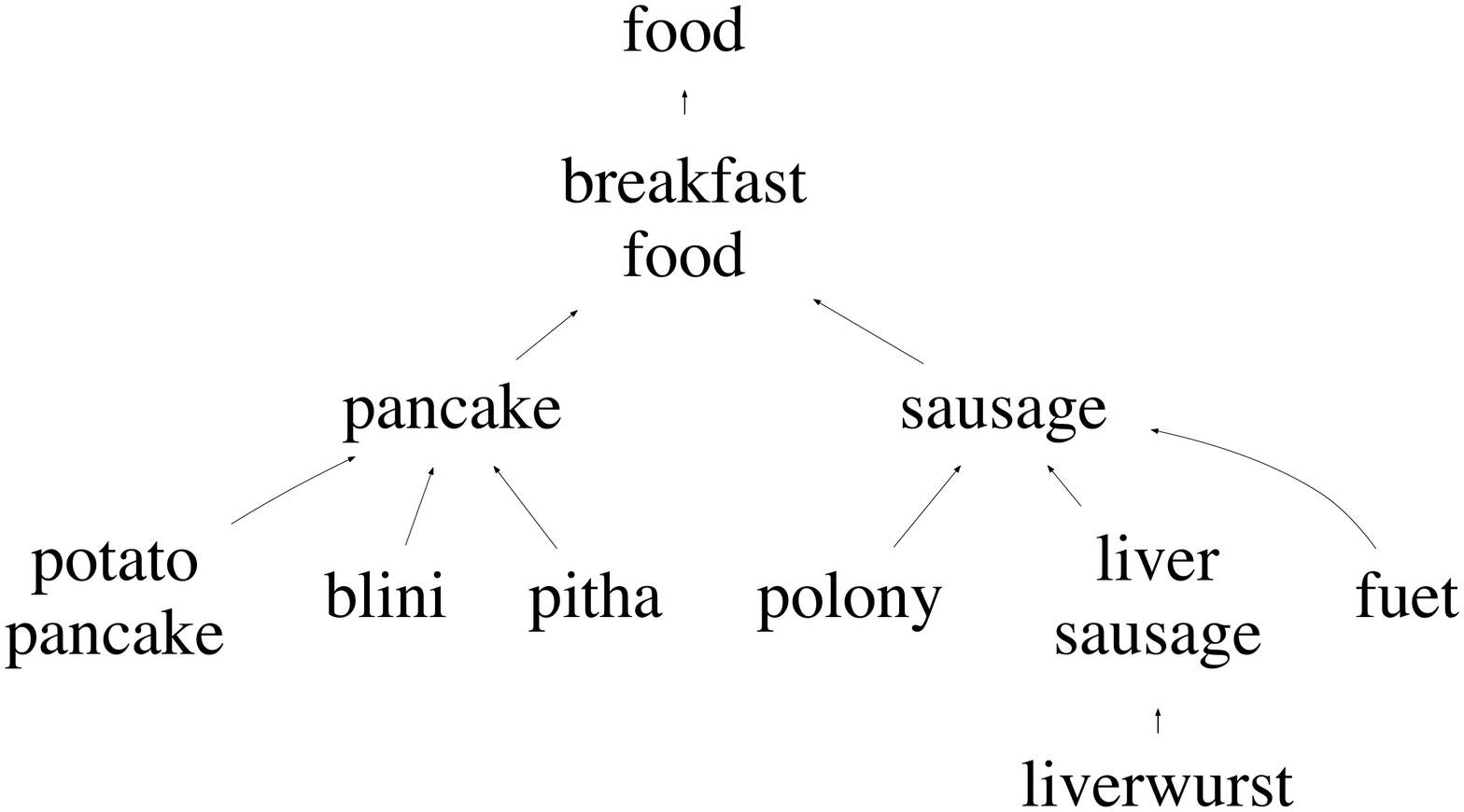}
        \caption{A section of SubSeq taxonomy ($\alpha$=$0.9$).}
        \label{fig:tax}
    \end{minipage}%
  \vspace{0.5cm}
   \begin{minipage}{0.3\textwidth}
        \centering
       \includegraphics[width=\linewidth]{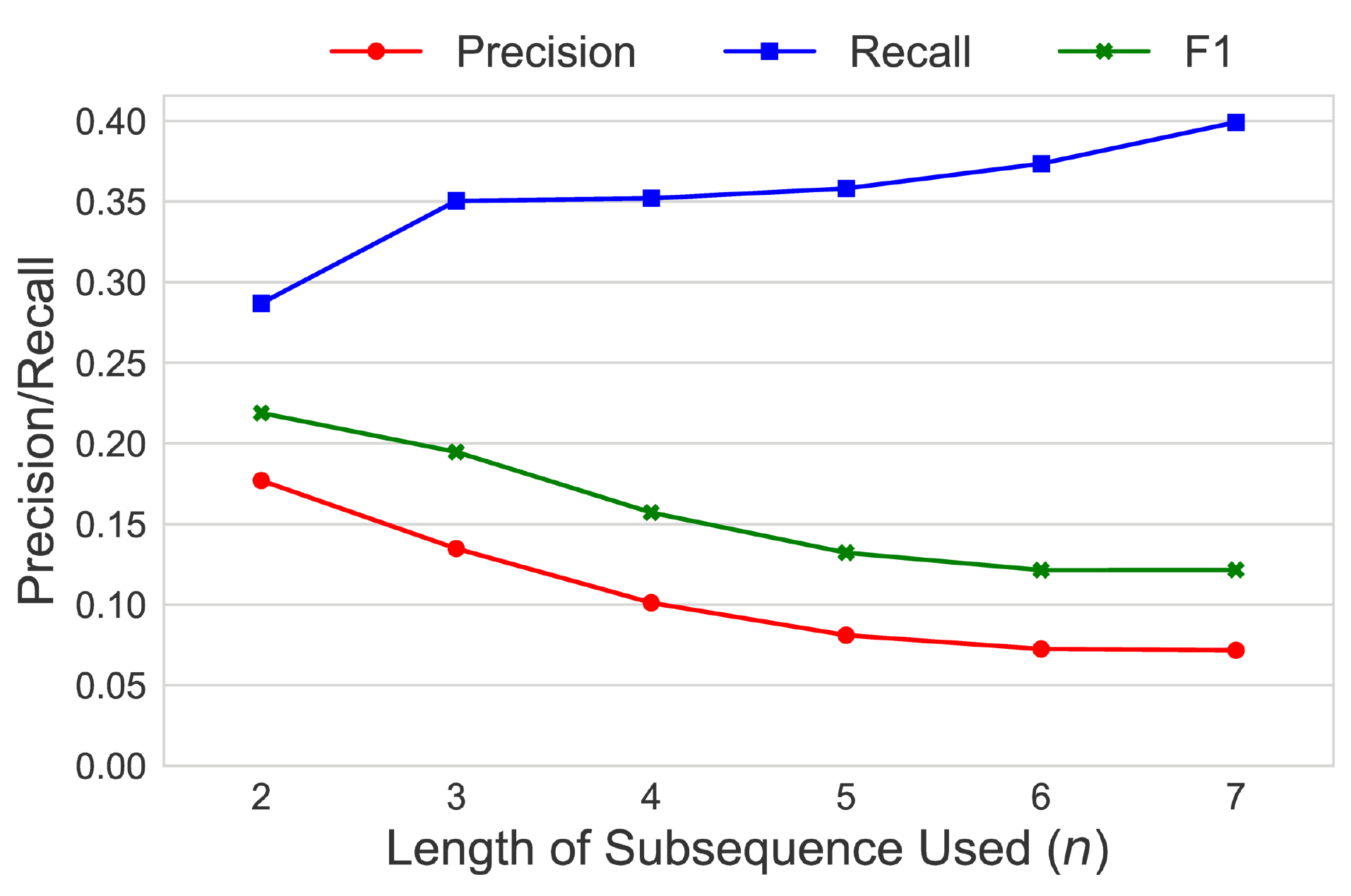}
        \caption{Precision/Recall vs. subsequence length ($n$).}
        \label{fig:prbysl}
    \end{minipage}%
    \quad
   \begin{minipage}{0.3\textwidth}
        \centering
       \includegraphics[width=\linewidth]{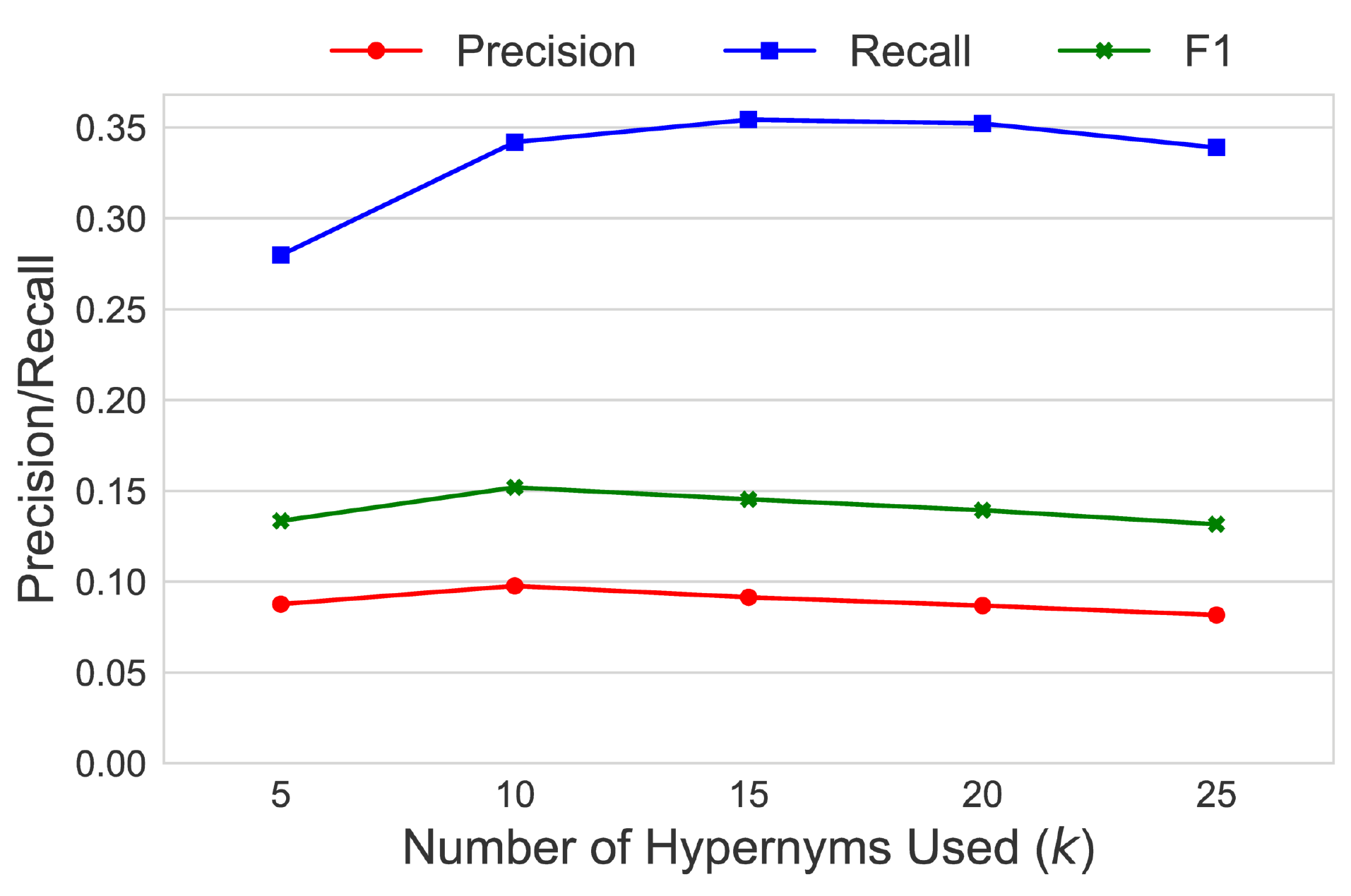}
        \caption{Precision/Recall vs. number of hypernyms used ($k$).}
        \label{fig:prbyk}
    \end{minipage}%
    \quad
    \begin{minipage}{0.3\textwidth}
        \centering
       \includegraphics[width=\linewidth]{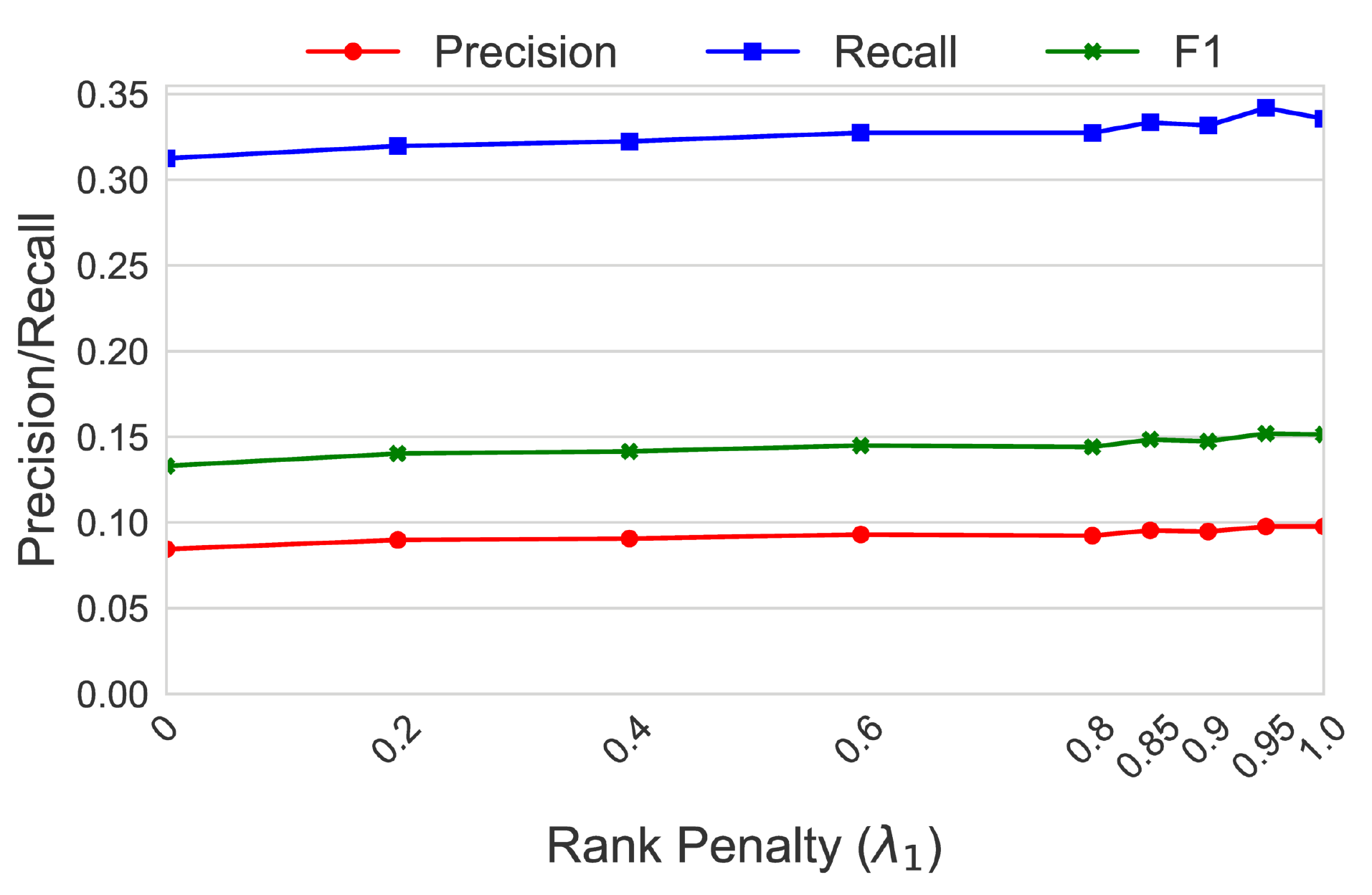}
        \caption{Precision/Recall vs. rank penalty (${\lambda}_1$).}
        \label{fig:prbyl}
    \end{minipage}%
    
    \quad
      
  \vspace{-0.25cm}
\end{figure*}

\subsection{Evaluation with Noisy Vocabulary}
\label{sec:noisy}

In the previous experiment, we performed taxonomy induction under the simplifying assumption that a clean input vocabulary of relevant domain terms is available. However, as explained in Section~\ref{sec:intro}, in practice, this assumption is rarely satisfied for most domains. Hence, in this experiment, we evaluate the performance of SubSeq in the presence of significant noise in the input vocabulary.

TAXI is inapplicable in this setting, as it assumes a clean input vocabulary consisting of both leaf and non-leaf terms. Instead, we compare SubSeq against a baseline, which is an edges-based variant of SubSeq.

\paragraph{\textbf{Setup.}} We first build a corpus of relevant documents for the food domain by collecting all English Wikipedia articles with titles matching at least one seed term (post lemmatization) in the SemEval food vocabulary. In total, 1,344 matching Wikipedia articles are found from the initial set of 1,555 seed terms. We run \textit{TermSuite}~\citep{cram2016termsuite}, a state-of-the-art 
term extraction approach to extract an initial terminology of 12,645 terms. All terms with occurrence counts $<5$ in the corpus are removed, thus resulting in a final terminology of 3,977 terms. The final terminology contains numerous noisy terms that are not food items, such as \textit{South Asia} and \textit{triangular}.

We now describe the edge-based baseline, hereafter referred to as \textit{TopEdge}, which extracts individual hypernym edges for terms in the vocabulary. TopEdge is identical to SubSeq, except that rather than extracting hypernym subsequences, it extracts direct hypernyms for terms with the highest hypernym probability $\text{Pr}_e(x_1,x_2)$ (cf. Equation~\ref{eqn:features}). It starts with the seed terms, and recursively extracts hypernyms for terms that do not yet have a hypernym until a fixed number of iterations. The aggregation and taxonomy construction steps are identical to SubSeq (cf. Sections~\ref{sec:agg} and~\ref{sec:flow}). Since the only difference between SubSeq and TopEdge is the extraction of hypernym subsequences compared to individual hypernym edges, this experiment also serves to evaluate the utility of extracting hypernym subsequences.

\paragraph{\textbf{Evaluation Results.}}We compare the quality of the taxonomies induced by TopEdge and SubSeq against the sub-hierarchy of WordNet rooted at \textit{food} as the gold standard. More specifically, we compute two metrics, i.e., \textit{term precision} and \textit{edge precision}. Term precision of a taxonomy is computed for the set of the input vocabulary terms retained by the taxonomy as: the ratio of the number of terms in the food sub-hierarchy of WordNet to the total number of terms present in WordNet.
Edge precision is computed as the ancestor precision: all nodes from the taxonomy that are not present in the WordNet are removed, and precision is computed on the hypernymy relations from the initial vocabulary to the root\footnote{Trivial edges $t\rightarrow$\textit{food} are ignored for all terms $t$.}. 

Figures~\ref{fig:alpha3} and~\ref{fig:alpha2} show the term precision and edge precision for TopEdge and SubSeq taxonomy induction methods for varying values of required coverage, i.e., $\alpha$ (cf. Section~\ref{sec:flow}). Both Term and edge precision scores for SubSeq are significantly higher than TopEdge across all values of $\alpha$, hence demonstrating the utility of hypernym subsequences. For both methods, precision scores decrease with increase in $\alpha$. This behavior is expected, because as $\alpha$ increases additional potentially-noisy seed terms are included in the output taxonomies. Figure~\ref{fig:tax} shows a section of the SubSeq taxonomy for $\alpha$=$0.9$.

We also performed a manual evaluation to judge the quality of the taxonomic edges that are \textit{not} present in the WordNet. Two authors independently annotated 100 such edges each of TopEdge and SubSeq taxonomies for $\alpha$$=$$0.5$. The precision for SubSeq was found to be 86\% compared to 52\% for TopEdge, with a high inter-annotator agreement (0.68). Both evaluations show that the precision of SubSeq taxonomies is quite high, thus demonstrating the efficacy of SubSeq in inducing taxonomies from noisy terminologies.

When $\alpha$$=$$1$, i.e., all input terms are included in the final taxonomy, term precision is 30\%, indicating that only 30\% of the terms extracted by the terminology extraction algorithm belong to the WordNet food sub-hierarchy. In contrast, the term precision for the original seed terms provided by SemEval is 75.8\%, hence confirming the presence of significant noise in the output of the terminology extraction approach.

 Overall, this experiment demonstrates that SubSeq is an effective approach towards taxonomy induction under the presence of significant noise in input terminologies. It also shows that extraction of hypernym subsequences is beneficial and results in significantly more accurate taxonomies.

\paragraph{\textbf{Parameter Sensitivity.}}
We now discuss the effect of parameters on the efficacy of subsequence extraction. To this end, we first construct a gold standard by sampling a set of 100 terms from the food domain randomly and extracting their generalization paths from WordNet. For a set of parameters, we run subsequence extraction and compute the precision and recall averaged over the top-5 paths per term.
The parameters we focus on are the: subsequence length ($n$), number of hypernyms used ($k$), and rank-penalty (${\lambda}_1$) (cf. Equations~\ref{eqn:4} and ~\ref{eqn:5}). 

Figure~\ref{fig:prbysl} shows the precision/recall values for varying values of subsequence lengths (before the expansion phase). Precision decreases and recall increases as the subsequence length increases. This can be intuitively explained by the observation that candidate hypernyms (cf. Table~\ref{tab:apple_hyp}) usually only contain hypernyms up to 3/4 levels. Hence, longer subsequences would typically drift from the original term, thus causing loss of precision. Figure~\ref{fig:prbyk} shows the effect of the number of candidate hypernyms used ($k$) for subsequence extraction. As $k$ increases, both precision and recall increase initially, but drop afterwards. This shows the benefit of utilizing lower-ranked hypernyms for subsequence extraction. However, it also illustrates the significant noise present in candidate hypernyms beyond a certain $k$. Figure~\ref{fig:prbyl} shows the effect of rank-penalty (${\lambda}_1$), the parameter used to penalize candidate hypernyms with lower frequency counts. Both precision and recall are low for lower values of ${\lambda}_1$ and peak at ${\lambda}_1$$=$$0.95$. 

We also evaluated the sensitivity to other parameters. We found out that subsequence extraction is fairly stable across different values of beam width and length penalty (${\lambda}_2$). Moreover, we observed that the number of subsequences per term ($b$ in Equation~\ref{eqn:4}) is also inconsequential beyond a value of $4$ as irrelevant subsequences are filtered out by domain filtering (cf. Section~\ref{sec:tax}).

%% file: related.tex
\section{Related Work}
\label{sec:related}

Taxonomy induction is a well-studied task, and multiple different lines of work have been proposed in the prior literature. Early work on taxonomy induction aims to extend the existing partial taxonomies (e.g., WordNet) by inserting missing terms at appropriate positions.~\citet{widdows2003unsupervised} places the missing terms in regions with most semantically-similar neighbors.~\citet{snow2006semantic} use a probabilistic model to attach novel terms in an incremental greedy fashion, such that the conditional probability of a set of relational evidence given a taxonomy is maximized.~\citet{yang2009metric} cluster terms incrementally using an ontology metric learnt from a set of heterogeneous features such as co-occurrence, context, and lexico-syntactic patterns.

A different line of work aims to exploit collaboratively-built semi-structured content such as Wikipedia for inducing large-scale taxonomies. Wikipedia links millions of entities (e.g., \textit{Johnny Depp}) to a network of inter-connected categories of different granularity (e.g. \textit{Hollywood Actors}, \textit{Celebrities}). WikiTaxonomy \cite{ponzetto2007deriving,ponzetto2008wikitaxonomy} labels these links as hypernymy or non-hypernymy, using a cascade of heuristics based on the syntactic structure of Wikipedia category labels, the topology of the network and lexico-syntactic patterns for detecting subsumption and meronymy, similar to Hearst patterns~\cite{hearst1992automatic}. 
WikiNet \cite{nastase2010wikinet} extends WikiTaxonomy by expanding non-hypernymy relations into fine-grained relations such as \textit{part-of, located-in, etc}. YAGO induces a taxonomy by employing heuristics linking Wikipedia categories to corresponding synsets in WordNet \cite{hoffart2013yago2}. More recently, ~\citet{flati2016multiwibi} and ~\citet{gupta2017280} propose approaches towards multilingual taxonomy induction from Wikipedia, resulting in taxonomies for over 270 languages. However, as pointed out by ~\citet{hovy2013collaboratively}, these taxonomy induction approaches are non-transferable, i.e., they only work for Wikipedia, because they employ lightweight heuristics that exploit the semi-structured nature of Wikipedia content. 

Although taxonomy induction approaches based on external lexical resources achieve high precision, they usually suffer from incomplete coverage over specific domains. To address  this issue, another line of work focuses on building lexical taxonomies automatically from a domain-specific corpus or Web.~\citet{kozareva2010semi} start from an initial set of root terms and basic level terms and use hearst-like lexico-syntactic patterns recursively to harvest new terms from the Web. Hypernymy relations between terms are induced by searching the Web again with surface patterns.
The graph of extracted hypernyms is subsequently pruned using heuristics based on the out-degree of nodes and the path lengths between terms.~\citet{velardi2013ontolearn} extract hypernymy relations from textual definitions discovered on the Web, and further employ an optimal branching algorithm to induce a taxonomy.

More recently,~\citet{task17semeval2015,task13semeval2016} introduced the first shared tasks on open-domain Taxonomy Extraction, thus providing a common ground for evaluation. INRIASAC, the top system in 2015 task, uses features based on substrings and co-occurrence statistics~\citep{grefenstette2015inriasac} whereas TAXI, the top system in 2016 task, uses lexico-syntactic patterns, substrings and focused crawling~\citep{panchenko2016taxi}.

In contrast to taxonomy induction approaches which use external resources, taxonomy induction approaches from a domain corpus or Web typically face two main obstacles. First, they assume the availability of a clean input vocabulary of seed terms. This requirement is not satisfied for most domains, thus requiring a time-consuming manual cleaning of noisy input vocabularies. 
Second, they ignore the relationship between terms and senses. For example, taxonomies induced from WordNet or Wikipedia produce different hypernyms for each sense of the term \textit{apple} (e.g., \textit{apple} is a \textit{fruit} or a \textit{company}). To tackle the second obstacle, taxonomy induction approaches from a domain corpus employ domain filtering to perform implicit sense disambiguation. This is done by removing hypernyms corresponding to domain-irrelevant senses of the terms~\cite{velardi2013ontolearn}. Although taxonomies should ideally contain senses rather
than terms, term taxonomies have shown significant efficacy in a variety of NLP tasks~\cite{biemann2005ontology,velardi2013ontolearn,bansal2014structured}.

To put it in context, our approach is similar to the previous attempts at inducing taxonomies without using external resources such as WordNet or Wikipedia. One key differentiator, however, is that it is robust to the presence of significant noise in the input vocabulary, thus dealing with the first obstacle above. To deal with the second obstacle, our approach performs implicit sense disambiguation via domain filtering at two different steps: (i) domain filtering of subsequences (cf. Section~\ref{sec:agg}); (ii) assigning lower cost for likely in-domain edges when applying the minimum-cost flow optimization (cf. Section ~\ref{sec:agg} \&~\ref{sec:flow}).

%% file: conclusions.tex
\section{Conclusions}
In this paper, we proposed a novel probabilistic framework for extracting hypernym subsequences from individual hypernymy relations. We also presented a minimum cost-flow optimization approach to taxonomy induction from a noisy hypernym graph. We demonstrated that our subsequence-based approach outperforms state-of-the-art taxonomy induction approaches that utilize individual hypernymy edge features.
Unlike previous approaches, our taxonomy induction approach is robust to the significant presence of noise in the input terminology. It also provides a user-defined parameter for controlling the accuracy and coverage of terms and edges in output taxonomies. As a consequence, our approach is applicable to arbitrary domains without any manual intervention, thus truly automating the process of taxonomy induction.

\label{sec:conc}

%% file: main.bbl

\begin{thebibliography}{00}


\ifx \showCODEN    \undefined \def \showCODEN     #1{\unskip}     \fi
\ifx \showDOI      \undefined \def \showDOI       #1{#1}\fi
\ifx \showISBNx    \undefined \def \showISBNx     #1{\unskip}     \fi
\ifx \showISBNxiii \undefined \def \showISBNxiii  #1{\unskip}     \fi
\ifx \showISSN     \undefined \def \showISSN      #1{\unskip}     \fi
\ifx \showLCCN     \undefined \def \showLCCN      #1{\unskip}     \fi
\ifx \shownote     \undefined \def \shownote      #1{#1}          \fi
\ifx \showarticletitle \undefined \def \showarticletitle #1{#1}   \fi
\ifx \showURL      \undefined \def \showURL       {\relax}        \fi
\providecommand\bibfield[2]{#2}
\providecommand\bibinfo[2]{#2}
\providecommand\natexlab[1]{#1}
\providecommand\showeprint[2][]{arXiv:#2}

\bibitem[\protect\citeauthoryear{Alfarone and Davis}{Alfarone and
  Davis}{2015}]%
        {alfarone2015unsupervised}
\bibfield{author}{\bibinfo{person}{Daniele Alfarone} {and}
  \bibinfo{person}{Jesse Davis}.} \bibinfo{year}{2015}\natexlab{}.
\newblock \showarticletitle{Unsupervised learning of an is-a taxonomy from a
  limited domain-specific corpus}. In \bibinfo{booktitle}{{\em Proceedings of
  the 24th International Joint Conference on Artificial Intelligence}}. AAAI
  Press, \bibinfo{pages}{1434--1441}.
\newblock


\bibitem[\protect\citeauthoryear{Bansal, Burkett, De~Melo, and Klein}{Bansal
  et~al\mbox{.}}{2014}]%
        {bansal2014structured}
\bibfield{author}{\bibinfo{person}{Mohit Bansal}, \bibinfo{person}{David
  Burkett}, \bibinfo{person}{Gerard De~Melo}, {and} \bibinfo{person}{Dan
  Klein}.} \bibinfo{year}{2014}\natexlab{}.
\newblock \showarticletitle{Structured Learning for Taxonomy Induction with
  Belief Propagation.}. In \bibinfo{booktitle}{{\em ACL (1)}}.
  \bibinfo{pages}{1041--1051}.
\newblock


\bibitem[\protect\citeauthoryear{Biemann}{Biemann}{2005}]%
        {biemann2005ontology}
\bibfield{author}{\bibinfo{person}{Chris Biemann}.}
  \bibinfo{year}{2005}\natexlab{}.
\newblock \showarticletitle{Ontology Learning from Text: {A} Survey of
  Methods}.
\newblock \bibinfo{journal}{{\em {LDV} Forum\/}} \bibinfo{volume}{20},
  \bibinfo{number}{2} (\bibinfo{year}{2005}), \bibinfo{pages}{75--93}.
\newblock
\showURL{%
\url{http://www.jlcl.org/2005_Heft2/Chris_Biemann.pdf}}


\bibitem[\protect\citeauthoryear{Bordea, Buitelaar, Faralli, and
  Navigli}{Bordea et~al\mbox{.}}{2015}]%
        {task17semeval2015}
\bibfield{author}{\bibinfo{person}{Georgeta Bordea}, \bibinfo{person}{Paul
  Buitelaar}, \bibinfo{person}{Stefano Faralli}, {and} \bibinfo{person}{Roberto
  Navigli}.} \bibinfo{year}{2015}\natexlab{}.
\newblock \showarticletitle{Semeval-2015 task 17: Taxonomy Extraction
  Evaluation (TExEval)}. In \bibinfo{booktitle}{{\em Proceedings of the 9th
  International Workshop on Semantic Evaluation}}. Association for
  Computational Linguistics.
\newblock


\bibitem[\protect\citeauthoryear{Bordea, Lefever, and Buitelaar}{Bordea
  et~al\mbox{.}}{2016}]%
        {task13semeval2016}
\bibfield{author}{\bibinfo{person}{Georgeta Bordea}, \bibinfo{person}{Els
  Lefever}, {and} \bibinfo{person}{Paul Buitelaar}.}
  \bibinfo{year}{2016}\natexlab{}.
\newblock \showarticletitle{Semeval-2016 task 13: Taxonomy Extraction
  Evaluation (TExEval-2)}. In \bibinfo{booktitle}{{\em Proceedings of the 10th
  International Workshop on Semantic Evaluation}}. Association for
  Computational Linguistics.
\newblock


\bibitem[\protect\citeauthoryear{Cram and Daille}{Cram and Daille}{2016}]%
        {cram2016termsuite}
\bibfield{author}{\bibinfo{person}{Damien Cram} {and}
  \bibinfo{person}{B{\'e}atrice Daille}.} \bibinfo{year}{2016}\natexlab{}.
\newblock \showarticletitle{Termsuite: Terminology extraction with term variant
  detection}.
\newblock \bibinfo{journal}{{\em ACL 2016\/}} (\bibinfo{year}{2016}),
  \bibinfo{pages}{13}.
\newblock


\bibitem[\protect\citeauthoryear{Flati, Vannella, Pasini, and Navigli}{Flati
  et~al\mbox{.}}{2016}]%
        {flati2016multiwibi}
\bibfield{author}{\bibinfo{person}{Tiziano Flati}, \bibinfo{person}{Daniele
  Vannella}, \bibinfo{person}{Tommaso Pasini}, {and} \bibinfo{person}{Roberto
  Navigli}.} \bibinfo{year}{2016}\natexlab{}.
\newblock \showarticletitle{MultiWiBi: The multilingual Wikipedia bitaxonomy
  project}.
\newblock \bibinfo{journal}{{\em Artif. Intell.\/}}  \bibinfo{volume}{241}
  (\bibinfo{year}{2016}), \bibinfo{pages}{66--102}.
\newblock
\showDOI{%
\url{https://doi.org/10.1016/j.artint.2016.08.004}}


\bibitem[\protect\citeauthoryear{Grefenstette}{Grefenstette}{2015}]%
        {grefenstette2015inriasac}
\bibfield{author}{\bibinfo{person}{Gregory Grefenstette}.}
  \bibinfo{year}{2015}\natexlab{}.
\newblock \showarticletitle{INRIASAC: Simple hypernym extraction methods}.
\newblock \bibinfo{journal}{{\em arXiv preprint arXiv:1502.01271\/}}
  (\bibinfo{year}{2015}).
\newblock


\bibitem[\protect\citeauthoryear{Gupta, Lebret, Harkous, and Aberer}{Gupta
  et~al\mbox{.}}{2017}]%
        {gupta2017280}
\bibfield{author}{\bibinfo{person}{Amit Gupta}, \bibinfo{person}{R{\'e}mi
  Lebret}, \bibinfo{person}{Hamza Harkous}, {and} \bibinfo{person}{Karl
  Aberer}.} \bibinfo{year}{2017}\natexlab{}.
\newblock \showarticletitle{280 Birds with One Stone: Inducing Multilingual
  Taxonomies from Wikipedia using Character-level Classification}.
\newblock \bibinfo{journal}{{\em arXiv preprint arXiv:1704.07624\/}}
  (\bibinfo{year}{2017}).
\newblock


\bibitem[\protect\citeauthoryear{Gupta, Piccinno, Kozhevnikov, Pasca, and
  Pighin}{Gupta et~al\mbox{.}}{2016}]%
        {guptarevisiting}
\bibfield{author}{\bibinfo{person}{Amit Gupta}, \bibinfo{person}{Francesco
  Piccinno}, \bibinfo{person}{Mikhail Kozhevnikov}, \bibinfo{person}{Marius
  Pasca}, {and} \bibinfo{person}{Daniele Pighin}.}
  \bibinfo{year}{2016}\natexlab{}.
\newblock \showarticletitle{Revisiting Taxonomy Induction over Wikipedia}. In
  \bibinfo{booktitle}{{\em {COLING} 2016, 26th International Conference on
  Computational Linguistics, Proceedings of the Conference: Technical Papers,
  December 11-16, 2016, Osaka, Japan}}. \bibinfo{pages}{2300--2309}.
\newblock


\bibitem[\protect\citeauthoryear{Gurevych and Wolf}{Gurevych and Wolf}{2010}]%
        {gurevych2010expert}
\bibfield{author}{\bibinfo{person}{Iryna Gurevych} {and}
  \bibinfo{person}{Elisabeth Wolf}.} \bibinfo{year}{2010}\natexlab{}.
\newblock \showarticletitle{Expert-Built and Collaboratively Constructed
  Lexical Semantic Resources}.
\newblock \bibinfo{journal}{{\em Language and Linguistics Compass\/}}
  \bibinfo{volume}{4}, \bibinfo{number}{11} (\bibinfo{year}{2010}),
  \bibinfo{pages}{1074--1090}.
\newblock


\bibitem[\protect\citeauthoryear{Harris}{Harris}{1954}]%
        {harris1954distributional}
\bibfield{author}{\bibinfo{person}{Zellig~S Harris}.}
  \bibinfo{year}{1954}\natexlab{}.
\newblock \showarticletitle{Distributional structure}.
\newblock \bibinfo{journal}{{\em Word\/}} \bibinfo{volume}{10},
  \bibinfo{number}{2-3} (\bibinfo{year}{1954}), \bibinfo{pages}{146--162}.
\newblock


\bibitem[\protect\citeauthoryear{Hearst}{Hearst}{1992}]%
        {hearst1992automatic}
\bibfield{author}{\bibinfo{person}{Marti~A Hearst}.}
  \bibinfo{year}{1992}\natexlab{}.
\newblock \showarticletitle{Automatic acquisition of hyponyms from large text
  corpora}. In \bibinfo{booktitle}{{\em Proceedings of the 14th conference on
  Computational linguistics-Volume 2}}. Association for Computational
  Linguistics, \bibinfo{pages}{539--545}.
\newblock


\bibitem[\protect\citeauthoryear{Hoffart, Suchanek, Berberich, and
  Weikum}{Hoffart et~al\mbox{.}}{2013}]%
        {hoffart2013yago2}
\bibfield{author}{\bibinfo{person}{Johannes Hoffart},
  \bibinfo{person}{Fabian~M. Suchanek}, \bibinfo{person}{Klaus Berberich},
  {and} \bibinfo{person}{Gerhard Weikum}.} \bibinfo{year}{2013}\natexlab{}.
\newblock \showarticletitle{{YAGO2:} {A} spatially and temporally enhanced
  knowledge base from Wikipedia}.
\newblock \bibinfo{journal}{{\em Artif. Intell.\/}}  \bibinfo{volume}{194}
  (\bibinfo{year}{2013}), \bibinfo{pages}{28--61}.
\newblock


\bibitem[\protect\citeauthoryear{Hovy, Kozareva, and Riloff}{Hovy
  et~al\mbox{.}}{2009}]%
        {hovy2009toward}
\bibfield{author}{\bibinfo{person}{Eduard Hovy}, \bibinfo{person}{Zornitsa
  Kozareva}, {and} \bibinfo{person}{Ellen Riloff}.}
  \bibinfo{year}{2009}\natexlab{}.
\newblock \showarticletitle{Toward completeness in concept extraction and
  classification}. In \bibinfo{booktitle}{{\em Proceedings of the 2009
  Conference on Empirical Methods in Natural Language Processing: Volume
  2-Volume 2}}. Association for Computational Linguistics,
  \bibinfo{pages}{948--957}.
\newblock


\bibitem[\protect\citeauthoryear{Hovy, Navigli, and Ponzetto}{Hovy
  et~al\mbox{.}}{2013}]%
        {hovy2013collaboratively}
\bibfield{author}{\bibinfo{person}{Eduard~H. Hovy}, \bibinfo{person}{Roberto
  Navigli}, {and} \bibinfo{person}{Simone~Paolo Ponzetto}.}
  \bibinfo{year}{2013}\natexlab{}.
\newblock \showarticletitle{Collaboratively built semi-structured content and
  Artificial Intelligence: The story so far}.
\newblock \bibinfo{journal}{{\em Artif. Intell.\/}}  \bibinfo{volume}{194}
  (\bibinfo{year}{2013}), \bibinfo{pages}{2--27}.
\newblock
\showDOI{%
\url{https://doi.org/10.1016/j.artint.2012.10.002}}


\bibitem[\protect\citeauthoryear{Klein}{Klein}{1967}]%
        {klein1967primal}
\bibfield{author}{\bibinfo{person}{Morton Klein}.}
  \bibinfo{year}{1967}\natexlab{}.
\newblock \showarticletitle{A primal method for minimal cost flows with
  applications to the assignment and transportation problems}.
\newblock \bibinfo{journal}{{\em Management Science\/}} \bibinfo{volume}{14},
  \bibinfo{number}{3} (\bibinfo{year}{1967}), \bibinfo{pages}{205--220}.
\newblock


\bibitem[\protect\citeauthoryear{Kliegr, Zeman, and Dojchinovski}{Kliegr
  et~al\mbox{.}}{2014}]%
        {kliegr2014linked}
\bibfield{author}{\bibinfo{person}{Tom{\'a}{\v{s}} Kliegr},
  \bibinfo{person}{V{\'a}clav Zeman}, {and} \bibinfo{person}{Milan
  Dojchinovski}.} \bibinfo{year}{2014}\natexlab{}.
\newblock \showarticletitle{Linked hypernyms dataset-generation framework and
  use cases}. In \bibinfo{booktitle}{{\em 3rd Workshop on Linked Data in
  Linguistics: Multilingual Knowledge Resources and Natural Language
  Processing}}.
\newblock


\bibitem[\protect\citeauthoryear{Kozareva and Hovy}{Kozareva and Hovy}{2010}]%
        {kozareva2010semi}
\bibfield{author}{\bibinfo{person}{Zornitsa Kozareva} {and}
  \bibinfo{person}{Eduard Hovy}.} \bibinfo{year}{2010}\natexlab{}.
\newblock \showarticletitle{A semi-supervised method to learn and construct
  taxonomies using the web}. In \bibinfo{booktitle}{{\em Proceedings of the
  2010 conference on empirical methods in natural language processing}}.
  Association for Computational Linguistics, \bibinfo{pages}{1110--1118}.
\newblock


\bibitem[\protect\citeauthoryear{Kozareva, Riloff, and Hovy}{Kozareva
  et~al\mbox{.}}{2008}]%
        {kozareva2008semantic}
\bibfield{author}{\bibinfo{person}{Zornitsa Kozareva}, \bibinfo{person}{Ellen
  Riloff}, {and} \bibinfo{person}{Eduard~H Hovy}.}
  \bibinfo{year}{2008}\natexlab{}.
\newblock \showarticletitle{Semantic Class Learning from the Web with Hyponym
  Pattern Linkage Graphs.}. In \bibinfo{booktitle}{{\em ACL}},
  Vol.~\bibinfo{volume}{8}. \bibinfo{pages}{1048--1056}.
\newblock


\bibitem[\protect\citeauthoryear{Lenat}{Lenat}{1995}]%
        {lenat1995cyc}
\bibfield{author}{\bibinfo{person}{Douglas~B Lenat}.}
  \bibinfo{year}{1995}\natexlab{}.
\newblock \showarticletitle{CYC: A large-scale investment in knowledge
  infrastructure}.
\newblock \bibinfo{journal}{{\it Commun. ACM}} \bibinfo{volume}{38},
  \bibinfo{number}{11} (\bibinfo{year}{1995}), \bibinfo{pages}{33--38}.
\newblock


\bibitem[\protect\citeauthoryear{Miller}{Miller}{1994}]%
        {miller1995wordnet}
\bibfield{author}{\bibinfo{person}{George~A. Miller}.}
  \bibinfo{year}{1994}\natexlab{}.
\newblock \showarticletitle{{WORDNET:} {A} Lexical Database for English}. In
  \bibinfo{booktitle}{{\em Human Language Technology, Proceedings of a Workshop
  held at Plainsboro, New Jerey, USA, March 8-11, 1994}}.
\newblock
\showURL{%
\url{http://aclweb.org/anthology/H/H94/H94-1111.pdf}}


\bibitem[\protect\citeauthoryear{Nakashole, Weikum, and Suchanek}{Nakashole
  et~al\mbox{.}}{2012}]%
        {nakashole2012patty}
\bibfield{author}{\bibinfo{person}{Ndapandula Nakashole},
  \bibinfo{person}{Gerhard Weikum}, {and} \bibinfo{person}{Fabian Suchanek}.}
  \bibinfo{year}{2012}\natexlab{}.
\newblock \showarticletitle{PATTY: a taxonomy of relational patterns with
  semantic types}. In \bibinfo{booktitle}{{\em Proceedings of the 2012 Joint
  Conference on Empirical Methods in Natural Language Processing and
  Computational Natural Language Learning}}. Association for Computational
  Linguistics, \bibinfo{pages}{1135--1145}.
\newblock


\bibitem[\protect\citeauthoryear{Nastase, Strube, Boerschinger, Zirn, and
  Elghafari}{Nastase et~al\mbox{.}}{2010}]%
        {nastase2010wikinet}
\bibfield{author}{\bibinfo{person}{Vivi Nastase}, \bibinfo{person}{Michael
  Strube}, \bibinfo{person}{Benjamin Boerschinger},
  \bibinfo{person}{C{\"{a}}cilia Zirn}, {and} \bibinfo{person}{Anas
  Elghafari}.} \bibinfo{year}{2010}\natexlab{}.
\newblock \showarticletitle{WikiNet: {A} Very Large Scale Multi-Lingual Concept
  Network}. In \bibinfo{booktitle}{{\em Proceedings of the International
  Conference on Language Resources and Evaluation, {LREC} 2010, 17-23 May 2010,
  Valletta, Malta}}.
\newblock


\bibitem[\protect\citeauthoryear{Navigli, Velardi, and Faralli}{Navigli
  et~al\mbox{.}}{2011}]%
        {navigli2011graph}
\bibfield{author}{\bibinfo{person}{Roberto Navigli}, \bibinfo{person}{Paola
  Velardi}, {and} \bibinfo{person}{Stefano Faralli}.}
  \bibinfo{year}{2011}\natexlab{}.
\newblock \showarticletitle{A graph-based algorithm for inducing lexical
  taxonomies from scratch}. In \bibinfo{booktitle}{{\em IJCAI}},
  Vol.~\bibinfo{volume}{2}. \bibinfo{pages}{2}.
\newblock


\bibitem[\protect\citeauthoryear{Oakes}{Oakes}{2005}]%
        {oakes2005using}
\bibfield{author}{\bibinfo{person}{Michael~P Oakes}.}
  \bibinfo{year}{2005}\natexlab{}.
\newblock \showarticletitle{Using Hearst's Rules for the Automatic Acquisition
  of Hyponyms for Mining a Pharmaceutical Corpus.}. In \bibinfo{booktitle}{{\em
  RANLP Text Mining Workshop}}, Vol.~\bibinfo{volume}{5}.
  \bibinfo{pages}{63--67}.
\newblock


\bibitem[\protect\citeauthoryear{Orlin}{Orlin}{1997}]%
        {orlin1997polynomial}
\bibfield{author}{\bibinfo{person}{James~B Orlin}.}
  \bibinfo{year}{1997}\natexlab{}.
\newblock \showarticletitle{A polynomial time primal network simplex algorithm
  for minimum cost flows}.
\newblock \bibinfo{journal}{{\em Mathematical Programming\/}}
  \bibinfo{volume}{78}, \bibinfo{number}{2} (\bibinfo{year}{1997}),
  \bibinfo{pages}{109--129}.
\newblock


\bibitem[\protect\citeauthoryear{Panchenko, Faralli, Ruppert, Remus, Naets,
  Fairon, Ponzetto, and Biemann}{Panchenko et~al\mbox{.}}{2016}]%
        {panchenko2016taxi}
\bibfield{author}{\bibinfo{person}{Alexander Panchenko},
  \bibinfo{person}{Stefano Faralli}, \bibinfo{person}{Eugen Ruppert},
  \bibinfo{person}{Steffen Remus}, \bibinfo{person}{Hubert Naets},
  \bibinfo{person}{C{\'e}drick Fairon}, \bibinfo{person}{Simone~Paolo
  Ponzetto}, {and} \bibinfo{person}{Chris Biemann}.}
  \bibinfo{year}{2016}\natexlab{}.
\newblock \showarticletitle{TAXI at SemEval-2016 Task 13: a taxonomy induction
  method based on lexico-syntactic patterns, substrings and focused crawling}.
\newblock \bibinfo{journal}{{\em Proceedings of SemEval\/}}
  (\bibinfo{year}{2016}), \bibinfo{pages}{1320--1327}.
\newblock


\bibitem[\protect\citeauthoryear{Ponzetto and Strube}{Ponzetto and
  Strube}{2007}]%
        {ponzetto2007deriving}
\bibfield{author}{\bibinfo{person}{S. Ponzetto} {and} \bibinfo{person}{M.
  Strube}.} \bibinfo{year}{2007}\natexlab{}.
\newblock \showarticletitle{Deriving a Large Scale Taxonomy from {W}ikipedia}.
  In \bibinfo{booktitle}{{\em Proceedings of the 22nd National Conference on
  Artificial Intelligence}}. \bibinfo{address}{Vancouver, British Columbia},
  \bibinfo{pages}{1440--1445}.
\newblock


\bibitem[\protect\citeauthoryear{Ponzetto and Strube}{Ponzetto and
  Strube}{2008}]%
        {ponzetto2008wikitaxonomy}
\bibfield{author}{\bibinfo{person}{Simone~Paolo Ponzetto} {and}
  \bibinfo{person}{Michael Strube}.} \bibinfo{year}{2008}\natexlab{}.
\newblock \showarticletitle{WikiTaxonomy: {A} Large Scale Knowledge Resource}.
  In \bibinfo{booktitle}{{\em {ECAI} 2008 - 18th European Conference on
  Artificial Intelligence, Patras, Greece, July 21-25, 2008, Proceedings}}.
  \bibinfo{pages}{751--752}.
\newblock


\bibitem[\protect\citeauthoryear{Ponzetto and Strube}{Ponzetto and
  Strube}{2011}]%
        {ponzetto2011taxonomy}
\bibfield{author}{\bibinfo{person}{Simone~Paolo Ponzetto} {and}
  \bibinfo{person}{Michael Strube}.} \bibinfo{year}{2011}\natexlab{}.
\newblock \showarticletitle{Taxonomy induction based on a collaboratively built
  knowledge repository}.
\newblock \bibinfo{journal}{{\em Artificial Intelligence\/}}
  \bibinfo{volume}{175}, \bibinfo{number}{9-10} (\bibinfo{year}{2011}),
  \bibinfo{pages}{1737--1756}.
\newblock


\bibitem[\protect\citeauthoryear{Sclano and Velardi}{Sclano and
  Velardi}{2007}]%
        {sclano2007termextractor}
\bibfield{author}{\bibinfo{person}{Francesco Sclano} {and}
  \bibinfo{person}{Paola Velardi}.} \bibinfo{year}{2007}\natexlab{}.
\newblock \showarticletitle{Termextractor: a web application to learn the
  shared terminology of emergent web communities}.
\newblock In \bibinfo{booktitle}{{\em Enterprise Interoperability II}}.
  \bibinfo{publisher}{Springer}, \bibinfo{pages}{287--290}.
\newblock


\bibitem[\protect\citeauthoryear{Seitner, Bizer, Eckert, Faralli, Meusel,
  Paulheim, and Ponzetto}{Seitner et~al\mbox{.}}{2016}]%
        {seitner2016large}
\bibfield{author}{\bibinfo{person}{Julian Seitner}, \bibinfo{person}{Christian
  Bizer}, \bibinfo{person}{Kai Eckert}, \bibinfo{person}{Stefano Faralli},
  \bibinfo{person}{Robert Meusel}, \bibinfo{person}{Heiko Paulheim}, {and}
  \bibinfo{person}{Simone~Paolo Ponzetto}.} \bibinfo{year}{2016}\natexlab{}.
\newblock \showarticletitle{A Large DataBase of Hypernymy Relations Extracted
  from the Web}. In \bibinfo{booktitle}{{\em Proceedings of the Tenth
  International Conference on Language Resources and Evaluation {LREC} 2016,
  Portoro{\v{z}}, Slovenia, May 23-28, 2016.}}
\newblock


\bibitem[\protect\citeauthoryear{Snow, Jurafsky, and Ng}{Snow
  et~al\mbox{.}}{2006}]%
        {snow2006semantic}
\bibfield{author}{\bibinfo{person}{Rion Snow}, \bibinfo{person}{Daniel
  Jurafsky}, {and} \bibinfo{person}{Andrew~Y Ng}.}
  \bibinfo{year}{2006}\natexlab{}.
\newblock \showarticletitle{Semantic taxonomy induction from heterogenous
  evidence}. In \bibinfo{booktitle}{{\em Proceedings of the 21st International
  Conference on Computational Linguistics and the 44th annual meeting of the
  Association for Computational Linguistics}}. Association for Computational
  Linguistics, \bibinfo{pages}{801--808}.
\newblock


\bibitem[\protect\citeauthoryear{Snow, Jurafsky, Ng, et~al\mbox{.}}{Snow
  et~al\mbox{.}}{2004}]%
        {snow2004learning}
\bibfield{author}{\bibinfo{person}{Rion Snow}, \bibinfo{person}{Daniel
  Jurafsky}, \bibinfo{person}{Andrew~Y Ng}, {and} \bibinfo{person}{others}.}
  \bibinfo{year}{2004}\natexlab{}.
\newblock \showarticletitle{Learning syntactic patterns for automatic hypernym
  discovery.}. In \bibinfo{booktitle}{{\em NIPS}}, Vol.~\bibinfo{volume}{17}.
  \bibinfo{pages}{1297--1304}.
\newblock


\bibitem[\protect\citeauthoryear{Suchanek, Kasneci, and Weikum}{Suchanek
  et~al\mbox{.}}{2007}]%
        {suchanek2007yago}
\bibfield{author}{\bibinfo{person}{Fabian~M. Suchanek},
  \bibinfo{person}{Gjergji Kasneci}, {and} \bibinfo{person}{Gerhard Weikum}.}
  \bibinfo{year}{2007}\natexlab{}.
\newblock \showarticletitle{Yago: a core of semantic knowledge}. In
  \bibinfo{booktitle}{{\em Proceedings of the 16th International Conference on
  World Wide Web, {WWW} 2007, Banff, Alberta, Canada, May 8-12, 2007}}.
  \bibinfo{pages}{697--706}.
\newblock


\bibitem[\protect\citeauthoryear{Suh, Convertino, Chi, and Pirolli}{Suh
  et~al\mbox{.}}{2009}]%
        {suh2009singularity}
\bibfield{author}{\bibinfo{person}{Bongwon Suh}, \bibinfo{person}{Gregorio
  Convertino}, \bibinfo{person}{Ed~H Chi}, {and} \bibinfo{person}{Peter
  Pirolli}.} \bibinfo{year}{2009}\natexlab{}.
\newblock \showarticletitle{The singularity is not near: slowing growth of
  Wikipedia}. In \bibinfo{booktitle}{{\em Proceedings of the 5th International
  Symposium on Wikis and Open Collaboration}}. ACM, \bibinfo{pages}{8}.
\newblock


\bibitem[\protect\citeauthoryear{Velardi, Faralli, and Navigli}{Velardi
  et~al\mbox{.}}{2013}]%
        {velardi2013ontolearn}
\bibfield{author}{\bibinfo{person}{Paola Velardi}, \bibinfo{person}{Stefano
  Faralli}, {and} \bibinfo{person}{Roberto Navigli}.}
  \bibinfo{year}{2013}\natexlab{}.
\newblock \showarticletitle{OntoLearn Reloaded: {A} Graph-Based Algorithm for
  Taxonomy Induction}.
\newblock \bibinfo{journal}{{\em Computational Linguistics\/}}
  \bibinfo{volume}{39}, \bibinfo{number}{3} (\bibinfo{year}{2013}),
  \bibinfo{pages}{665--707}.
\newblock
\showDOI{%
\url{https://doi.org/10.1162/COLI_a_00146}}


\bibitem[\protect\citeauthoryear{Widdows}{Widdows}{2003}]%
        {widdows2003unsupervised}
\bibfield{author}{\bibinfo{person}{Dominic Widdows}.}
  \bibinfo{year}{2003}\natexlab{}.
\newblock \showarticletitle{Unsupervised methods for developing taxonomies by
  combining syntactic and statistical information}. In \bibinfo{booktitle}{{\em
  Proceedings of the 2003 Conference of the North American Chapter of the
  Association for Computational Linguistics on Human Language Technology-Volume
  1}}. Association for Computational Linguistics, \bibinfo{pages}{197--204}.
\newblock


\bibitem[\protect\citeauthoryear{Yang and Callan}{Yang and Callan}{2009}]%
        {yang2009metric}
\bibfield{author}{\bibinfo{person}{Hui Yang} {and} \bibinfo{person}{Jamie
  Callan}.} \bibinfo{year}{2009}\natexlab{}.
\newblock \showarticletitle{A metric-based framework for automatic taxonomy
  induction}. In \bibinfo{booktitle}{{\em Proceedings of the Joint Conference
  of the 47th Annual Meeting of the ACL and the 4th International Joint
  Conference on Natural Language Processing of the AFNLP: Volume 1-Volume 1}}.
  Association for Computational Linguistics, \bibinfo{pages}{271--279}.
\newblock


\end{thebibliography}
